\documentclass[10pt,twocolumn,letterpaper]{article}
\usepackage{iccv}
\usepackage{times}
\usepackage{epsfig}
\usepackage{graphicx}
\usepackage{amsmath}
\usepackage{amssymb}
\usepackage{booktabs}
\usepackage{color}
\usepackage{bm}
\usepackage[pagebackref=true,breaklinks=true,letterpaper=true,colorlinks,bookmarks=false]{hyperref}
\iccvfinalcopy %

\renewcommand{\vec}[1]{\bm{#1}}
\newcommand\inputsingle{\vec{y}}
\newcommand\dinputsingle{\tilde{\inputsingle}}

\newcommand\poolvec{\vec{z}}

\newcommand\vectorize{\text{v}}

\newcommand{\trace}{\mathrm{tr}}

\renewcommand\paragraph[1]{\vspace{1pt}\noindent\textbf{#1}\quad}

\usepackage{verbatim}
\definecolor{light-gray}{gray}{0.5}

\begin{document}
\title{Generalized orderless pooling performs implicit salient matching}
\author{Marcel Simon$^1$, Yang Gao$^2$, Trevor Darrell$^2$, Joachim Denzler$^1$, Erik Rodner$^3$\\
$^1$ Computer Vision Group, University of Jena, Germany \hspace{0.5cm} $^2$ EECS, UC Berkeley, USA \\
$^3$ Corporate Research and Technology, Carl Zeiss AG\\
{\tt\small \{marcel.simon, joachim.denzler\}@uni-jena.de \enspace \tt\small\{yg, trevor\}@eecs.berkeley.edu}
}
\maketitle
\begin{abstract}

Most recent CNN architectures use average pooling as a final feature encoding step. 
In the field of fine-grained recognition, however, recent global representations like bilinear pooling offer improved performance.
In this paper, we generalize average and bilinear pooling to  ``$\alpha$-pooling'', allowing for learning the pooling strategy during training.
In addition, we present a novel way to visualize decisions made by these approaches. 
We identify parts of training images having the highest influence on the prediction of a given test image. 
It allows for justifying decisions to users and also for analyzing the influence of semantic parts.
For example, we can show that the higher capacity VGG16 model focuses much more on the bird's head than, e.g., the lower-capacity VGG-M model when recognizing fine-grained bird categories.
Both contributions allow us to analyze the difference when moving between average and bilinear pooling.
In addition, experiments show that our generalized approach can outperform both across a variety of standard datasets.
\end{abstract}
\section{Introduction}

\begin{figure}
  \centering
  \includegraphics[width=0.9\linewidth]{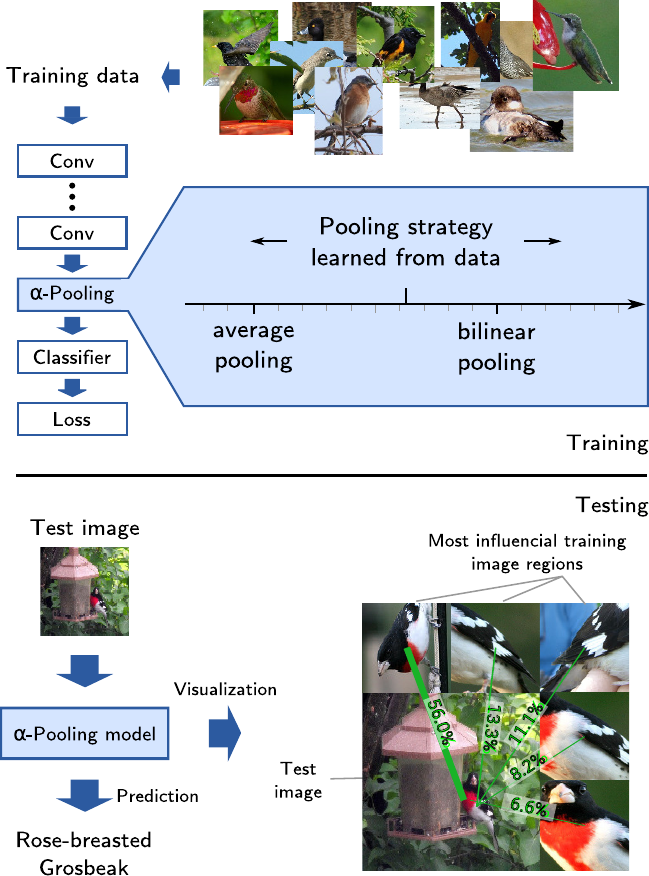}
  \vspace{5pt}
  \caption{We present the novel pooling strategy $\alpha$-pooling, which replaces the final average pooling or bilinear pooling layer in CNNs.
  It allows for a smooth combination of average and bilinear pooling techniques.
  The optimal pooling strategy can be learned during training to optimally adapt to the properties of the task.
  In addition, we present a novel way to visualize predictions of $\alpha$-pooling-based classification decisions. 
  It allows in particular for analyzing incorrect classification decisions, which is an important addition to all widely used orderless pooling strategies.
  }
  \label{fig:teaser}
\end{figure}

Deep architectures are characterized by interleaved convolution layers to compute intermediate features and pooling layers to aggregate information. Inspired by recent results in fine-grained recognition~\cite{lin15bilinear,gao15compactbilinear} showing certain pooling strategies offered equivalent performance as classic models involving explicit correspondence, we investigate here a new pooling layer generalization for deep neural networks suitable both for fine-grained and more generic recognition tasks.

Fine-grained recognition developed from a niche research field into a popular topic with numerous applications, ranging from automated monitoring of animal species~\cite{Freytag16_CFW} to fine-grained recognition of cloth types~\cite{di2013style}.
The defining property of fine-grained recognition is that all possible object categories share a similar object structure and hence similar object parts. 
Since the objects do not significantly differ in the overall shape, subtle differences in the appearance of an object part can likely make the difference between two classes.
For example, one of the most popular fine-grained tasks is bird species recognition.
All birds have the basic body structure with beak, head, throat, belly, wings as well as tail parts, and two species might differ only in the presence or absence of a yellow shade around the eyes.

Most approaches in the past five years concentrated on exploiting this extra knowledge about the shared general object structure.
Usually, the objects are described by the appearance of different parts.
This explicit modeling of the appearance of object parts is intuitive and natural.
While explicit part modeling greatly outperformed off-the-shelf CNNs, the recently presented %
second-order or bilinear pooling~\cite{lin15bilinear} %
gives a similar performance boost at the expense of an understandable model.

Our paper presents a novel approach which has \textit{both} state-of-the-art performance and  allows for clear justification of the classification prediction using visualization of influential training image regions.
Classification accuracy on this task is reaching human level performance and hence we additionally focus on making classification decisions more understandable and explainable.
We present an %
approach which can show for each evaluated image why the decision was made by referring to the most influential training image regions.
Average pooling is mainly used in generic recognition. while
bilinear pooling has its largest benefits in fine-grained recognition:
our approach allows for understanding and generalizing the relationship between the two -- a crucial step for further research.

Our \textit{first contribution} is a novel generalization and parametric representation of the commonly used average and bilinear pooling.
This representation allows for a smooth combination of these first-order and second-order operators.  This framework both provides a novel conceptual understanding of the relationship of the methods and offers a new operating point with a consistent improvement in terms of accuracy.

The \textit{second contribution} is an %
analysis of the learned optimal pooling strategy during training. 
Our parametrized pooling scheme is differentiable and hence can be integrated into an end-to-end-learnable pipeline. 
We show that the learned pooling scheme is related to the classification task it is trained on. 
In addition, a pooling scheme half-way between average and bilinear pooling seems to achieve the highest accuracy on several benchmark datasets.

Our \textit{third contribution} is a novel way to visually justify a classification prediction of a specific image to a user. 
It is complementary to our novel pooling scheme and hence also applicable to the previous pooling schemes average and bilinear pooling.
Both classifier parameters and local feature matches are considered to identify training image regions of highest influence.

Finally, our \textit{fourth contribution} is an approach for quantifying the influence of semantic parts in a classification decision. 
In contrast to previous work, we consider both the classifier parameters and the saliency.
We show that the human way of classifying objects, \ie using a broad set of object attributes, increasingly diverges from the CNN's way, which bases most of its decision on one object part.
In particular, we show that more complex CNN models like VGG16 focus much more on the bird's head compared to less complex ones like VGG-M.
We also show that a similar shift can be seen when moving from average pooled features to bilinear features encoding.

After reviewing related work in the following Section,
Section~\ref{sec:generalizing} will present our novel $\alpha$-pooling formulation, which generalizes average and bilinear pooling into a single framework.
Section~\ref{sec:pairwise-matching} will then investigate the relationship between generalized orderless pooling and pairwise matching and present an approach for visualizing a classification decision.
This is followed by the experiments and a discussion about the trade-offs between implicit and explicit pose normalization for fine-grained recognition
in Sec.~\ref{sec:experiments} and \ref{sec:discussion}.

\section{Related work}
Our work is related to several topics in the area of computer vision. 
This includes pooling techniques, match kernels, bilinear encoding, and visualizations for CNNs.

\paragraph{Pooling techniques and match kernels}
The presented $\alpha$-pooling is related to other pooling techniques, which aggregate a set of local features into a single feature vector.
Besides the commonly used average pooling, fully-connected layers, and maximum pooling, several new approaches have been developed in the last years.
Zeiler~\etal~\cite{zeiler13stochPooling} randomly pick in each channel an element according to a multinomial distribution, which is defined by the activations themselves.
Motivated by their success with hand-crafted features, Fisher vector \cite{gosselin14fvcnn,lin15bilinear} and VLAD encoding~\cite{gong14vladcnn} applied on top of the last convolutional layer have been evaluated as well.
The idea of spatial pyramids was used by He~\etal~\cite{he14spp} in order to improve recognition performance.
In contrast to these techniques, feature encoding based on $\alpha$-pooling show a significantly higher accuracy in fine-grained applications.
Lin~\etal~\cite{lin15bilinear,lin16texture} presents bilinear pooling, which is a special case of average pooling. 
It has its largest benefits in fine-grained tasks. 
As shown in the experiments, learning the right mix of average and bilinear pooling improves results especially in tasks besides fine-grained.

The relationship of average pooling and pairwise matching of local features was presented by Bo~\etal~\cite{bo09matchkernel} as an efficient encoding for matching a set of local features.  
This formulation was also briefly discussed in \cite{gao15compactbilinear} and used for deriving an explicit feature transformation, which approximates bilinear pooling.
Bilinear encoding was first mentioned by Tenenbaum~\etal~\cite{tenenbaum00bilinear} and used, for example, by Carreira~\etal~\cite{carreira12bilinear} and Lin~\etal~\cite{lin15bilinear} for image recognition tasks.
Furthermore, the recent work of Murray~\etal~\cite{murray2016interferences} also analyzes orderless pooling approaches and proposes a technique to
normalize the contribution of each local descriptor to resulting kernel values. In contrast, we show how the individual contributions can be used either for visualizing the classification decisions and for understanding the differences between generic and fine-grained tasks.

\paragraph{Justifying classifier predictions for an image}
Especially  Section~\ref{sec:pairwise-matching} is related to visualization techniques for information processing in CNNs.
Most of the previous works focused on the primal view of the feature representation.
This means they analyze the feature representations by looking only at a single image.
Zeiler~\etal~\cite{zeiler14visualizing} identify image patterns, which cause high activations of selected channels of a convolutional layer.
Yosinksi~\etal~\cite{yosinski15visualizing} try to generate input patterns, which lead to a maximal activation of certain units.
Bach~\etal~\cite{bach2015relevancePropagation} visualize areas important to the classification decision with layer-wise relevance propagation.
In contrast to the majority of these works, we focus on the dual (or kernel) view of image classification.
While a visualization for a single image looks interesting at the first sight, it does not allow for understanding which parts of an image are compared with which parts of the training images. 
In other words, these visualizations look only at the image itself and are omitting the relationship to the training data.
For example, while the bird's head might be an attentive region in the visualization techniques mentioned above, a system might still compare this head with some unrelated areas in other images. 
Our approach allows for a clearer understanding about which pairs of training and test image regions contribute to a classification decision.

Zhang~\etal~\cite{zhang16weakly} present an idea to realize this for the case of explicit part detectors. 
They use the prediction score of a SVM classifier for each part to identify the most important patches for a selected part detector.
We extend this idea to orderless-pooled features which do not originate from explicit part detections.

\section{From generic to fine-grained classification:
generalized $\alpha$-pooling \label{sec:generalizing}}

Fine-grained applications like bird recognition and more generic image classification tasks like ImageNet have traditionally been two related but clearly separate fields with their own specialized approaches. 
While the general CNN architecture is shared, its usage differs.
In this work, we focus on two state-of-the-art feature encoding: global average and bilinear pooling.
While bilinear pooling shows the largest benefits in fine-grained applications, average pooling is the most commonly chosen final pooling step in literally all state-of-the-art CNN architectures.
In this section, we show the connection between these two worlds.
We present the novel generalization $\alpha$-pooling, which allows for a continuous transition between average and bilinear pooling. 
The right mixture is learned with back-propagation from data in training, which allows for adapting the specific tasks.
In addition, the results will allow us to investigate which mixture of pooling approaches is best suited for which application, and what makes fine-grained recognition different from generic image classification.

\paragraph{Generalized $\bm{\alpha}$-pooling}
We propose a novel generalization of the common average and bilinear pooling as used in deep networks, which we call $\alpha$-pooling.
Let $(f, g, \mathcal{C})$ denote a classification model.
$f:\mathcal{I}\times i \mapsto \bm{y}_i \in \mathbb{R}^D$ denotes a local feature descriptor mapping from input image $\mathcal{I}$ and location $i$ to a vector with length $D$, which describes this region.
$g: \left\{ \bm{y}_i \,| \, i=1,\dots,n\right\} \mapsto \bm{z} \in \mathbb{R}^{M} $ is a pooling scheme which aggregates $n$ local features to a single global image description of length $M$.
In our case, $M=D^2$ and is compressed using \cite{gao15compactbilinear}.
Finally, $\mathcal{C}$ is a classifier.
In a common CNN like VGG16, $f$ corresponds to the first part of a CNN up to the last convolutional layer, $g$ are two fully connected layers and $\mathcal{C}$ is the final classifier.

An $\alpha$-pooling-model is then defined by $(f, g^\text{alpha}, \mathcal{C})$, where 
\begin{equation}
 g^\text{alpha}(\left\{ \bm{y}_i \right\}_{i=1}^n) = \vectorize \left(\frac{1}{n} \sum _{i=1}^n \text{alpha-prod}(\bm{y}_i, \alpha)\right)
\end{equation}
and
\begin{equation}
  \text{alpha-prod}(\bm{y}_i, \alpha) = (\text{sgn}(\bm{y}_i) \circ |\bm{y}_i|^{\alpha-1}) \bm{y}_i^T\,.
\end{equation}
where $\vectorize(\cdot)$ is the vectorization function, and $\text{sgn}(\cdot )$, $\cdot \circ \cdot$, $|\cdot|$, and $\cdot^\alpha$ denote the element-wise signum, product, absolute value and exponentiation function, and $\alpha$ is a model parameter. 
$\alpha$ has a significant influence on the pooling due to its role as an exponent.
The optimal value is learned with back-propagation. %
For numerical stability, we add a small constant $\epsilon>0$ to $|\inputsingle_i|$ when calculating the power and when calculating the logarithm.
In our experiments, learning $\alpha$ was stable.

\paragraph{Special cases}
Average pooling is a common final feature encoding step in most state-of-the-art CNN architectures like ResNet~\cite{he15resnet} or Inception~\cite{szegedy16inception4}. 
The combination~\cite{lin15bilinear} of CNN feature maps and bilinear pooling~\cite{tenenbaum00bilinear,carreira12bilinear} is one of the current state-of-the-art approaches in the fine-grained area.
Both approaches are a special case of $\alpha$-pooling.

For $\alpha=1$ and $\bm{y}\geq 0$, \ie $\text{alpha-prod}(\bm{y}_i, 1)=I \cdot \inputsingle_i^T$, $g^\text{alpha}$ calculates a matrix in which each row is the mean vector.
This mean vector is identical to the one obtained in common average pooling.
The vectorization $\vectorize(\cdot)$ turns the resulting matrix into a concatention of identical mean vectors.

In case of $\alpha=2$, \ie $\text{alpha-prod}(\bm{y}_i, 2)=\inputsingle_i\inputsingle_i^T$, the mean outer product of $\inputsingle_i$ is calculated, which is equivalent to bilinear pooling.
Therefore, $\alpha$-pooling allows for estimating the type of pooling necessary for a particular task by learning $\alpha$ directly from data.

$\alpha$-pooling can continuously shift between average and bilinear pooling, which opens a great variety of opportunities.
It shows a connection between both that was to the best of our knowledge previously unknown. 
Furthermore, and even more important, all following contributions are also applicable to these two commonly used pooling techniques.
They allow for analyzing and understanding differences between both special cases.

\section{Understanding decisions of $\alpha$-pooling \label{sec:pairwise-matching}}

In this section, we give a ``deep'' insight into the class of $\alpha$-pooled features, which includes average and bilinear pooling as well as shown in the last section. 
We make use of the formulation as pairwise matching of local features,
which allows for visualizing both the gist of the representation and resulting classification decisions.
We use the techniques presented in this section to analyze the effects of $\alpha$-pooled features as we move between generic and fine-grained classification tasks.
To simplify the notation, we will focus in this section on the case that all local features $\bm{y}$ are non-negative.
This is the case for all features used in the experiments.
All observations apply to the generic case in an analogous manner.

\paragraph{Interpreting decisions using most influential regions}
While an impressive classification accuracy can be achieved with orderless pooling, 
one of its main drawbacks is the difficulty interpreting classification decisions.
This applies especially to fine-grained tasks,
since the difference between two categories might not be clear even for a human expert.
Furthermore, there is a need to analyze false automatic predictions, to understand
error cases and advance algorithms.

In this section, we use the formulation of $\alpha$-pooling as pairwise matching to visualize classification decisions.
It is based on finding locations with high influence on the decision.
We show how to find the most relevant training image regions and show that even implicit part modeling approaches are well suited for visualizing decisions.

Let $\tilde{\bm{z}}$ be the $\alpha$-pooling representation of a new test image resulting
from local activations $\tilde{\inputsingle}_i$ of a convolutional layer.
If we use a single fully-connected layer after bilinear pooling and a suitable loss, the resulting
score for a single class is given up to a constant by the representer theorem as:
\begin{align}
    \sum\limits_{k=1}^N \beta_{k} \langle \bm{z}_k, \tilde{\bm{z}} \rangle
  = \sum\limits_{k=1}^N \sum\limits_{i,j} \beta_k \cdot \langle \inputsingle_{i,k}, \tilde{\inputsingle}_j \rangle \langle \inputsingle_{i,k}^{\alpha-1}, \tilde{\inputsingle}_j^{\alpha-1} \rangle,
\end{align}
where $\beta_k$ are the weights of each training image given by the dual representation
of the last layer and $N$ is the number of training samples.
$\bm{z}_k$ is the $\alpha$-pooled feature of the $k$-th training image and calculated using the local features $\inputsingle_{i,k}$.

A match between a region $j$ in the test example and region $i$ of a training example $k$ is defined
by the triplet $(k, i, j)$. The influence of the triplet on the final score is given
by the product
\begin{equation}
 \gamma_{k,i,j} = \beta_k \cdot \langle \inputsingle_{i,k}, \tilde{\inputsingle}_j \rangle \langle \inputsingle_{i,k}^{\alpha-1}, \tilde{\inputsingle}_j^{\alpha-1} \rangle \,. \label{eq:region-influence}
\end{equation}
Therefore, we can visualize the regions with the highest influence on the classification decisions
by showing the ones with the highest corresponding $\gamma_{k,i,j}$.
This calculation can be done efficiently also on large datasets with the main limitation being the memory for storing the feature maps.

\begin{figure}
 \centering
 Correct
 
 \includegraphics[width=0.49\linewidth]{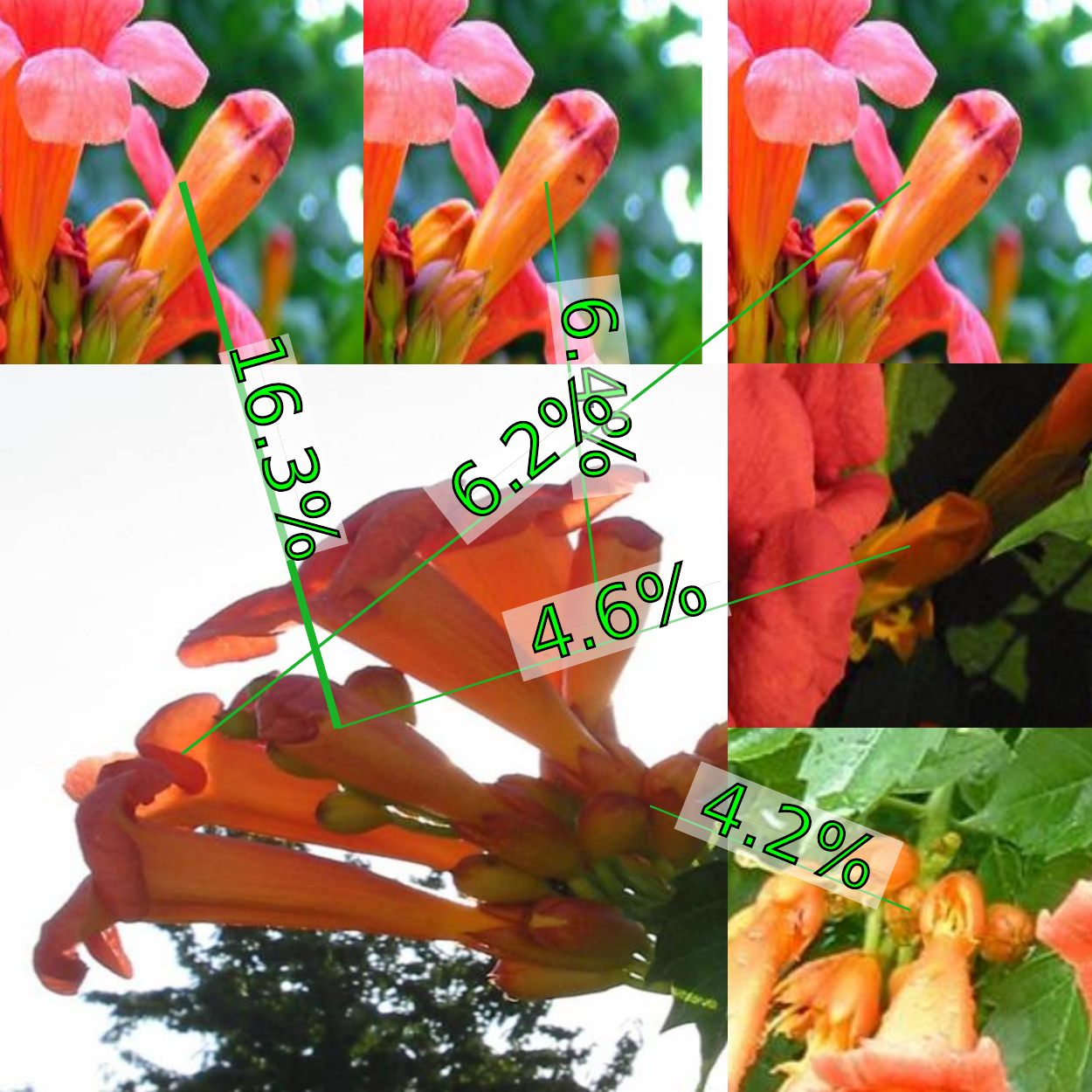}
 \includegraphics[width=0.49\linewidth]{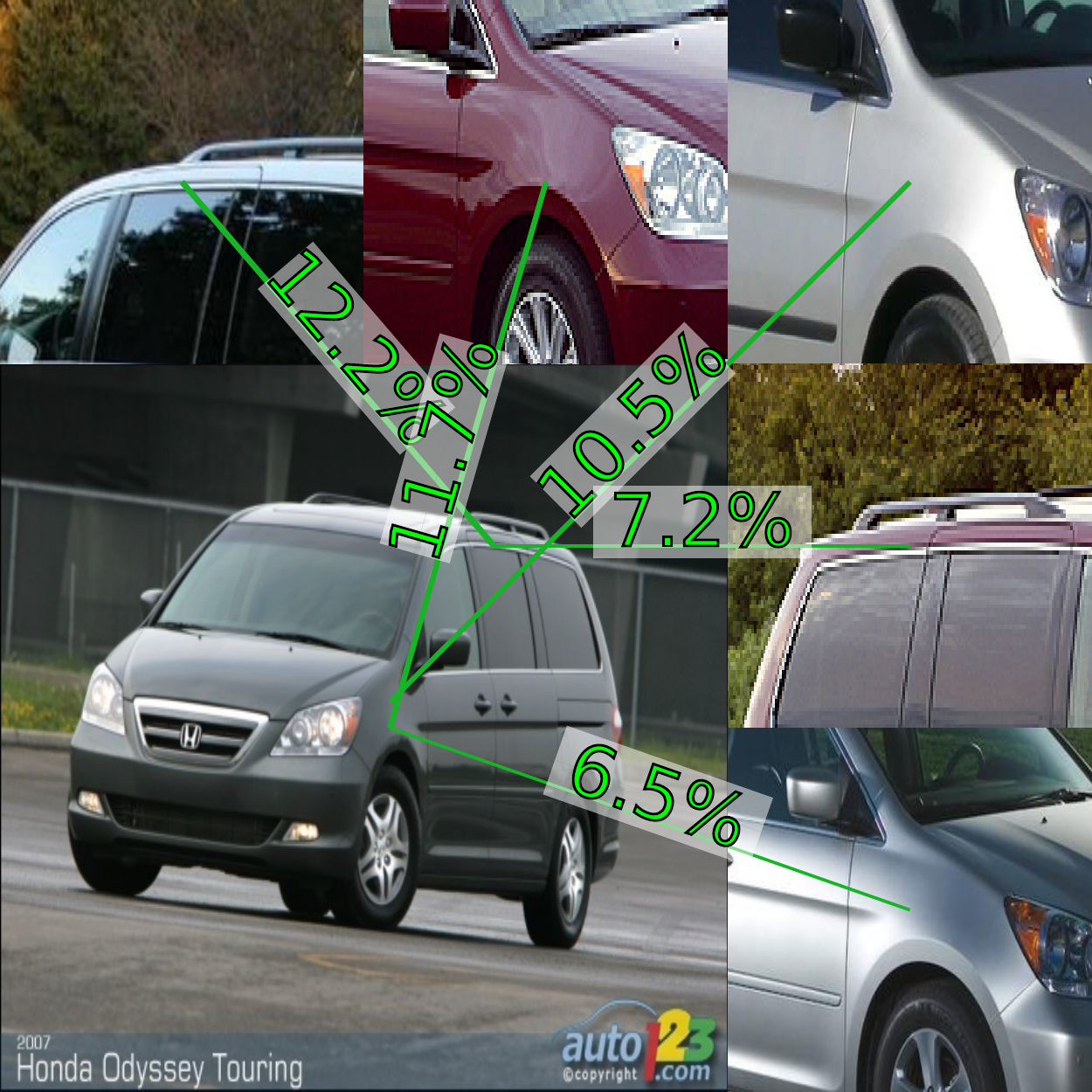}
 
 \vspace{2pt}
 \includegraphics[width=0.49\linewidth]{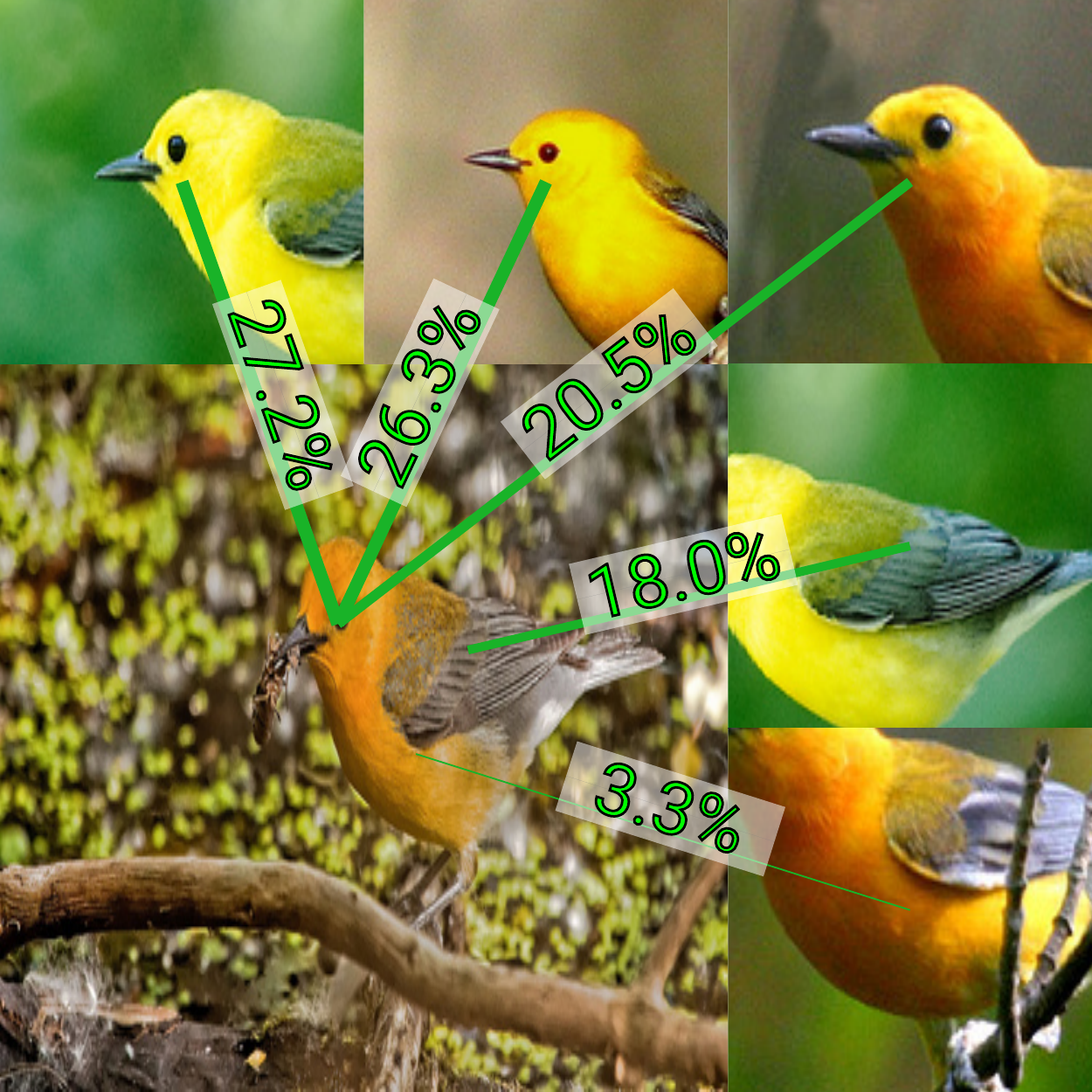}
 \includegraphics[width=0.49\linewidth]{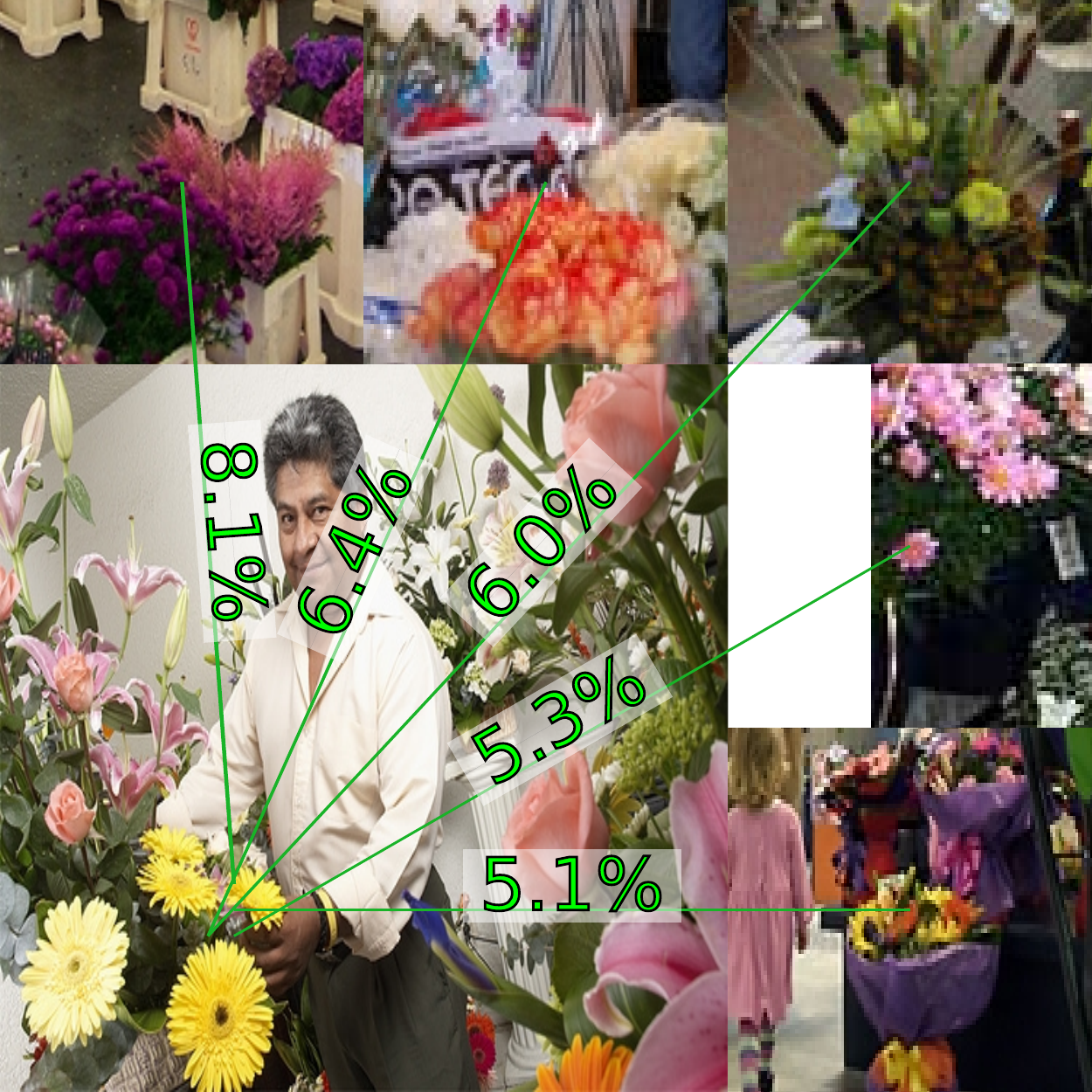}

 Incorrect
 
 \includegraphics[width=0.49\linewidth]{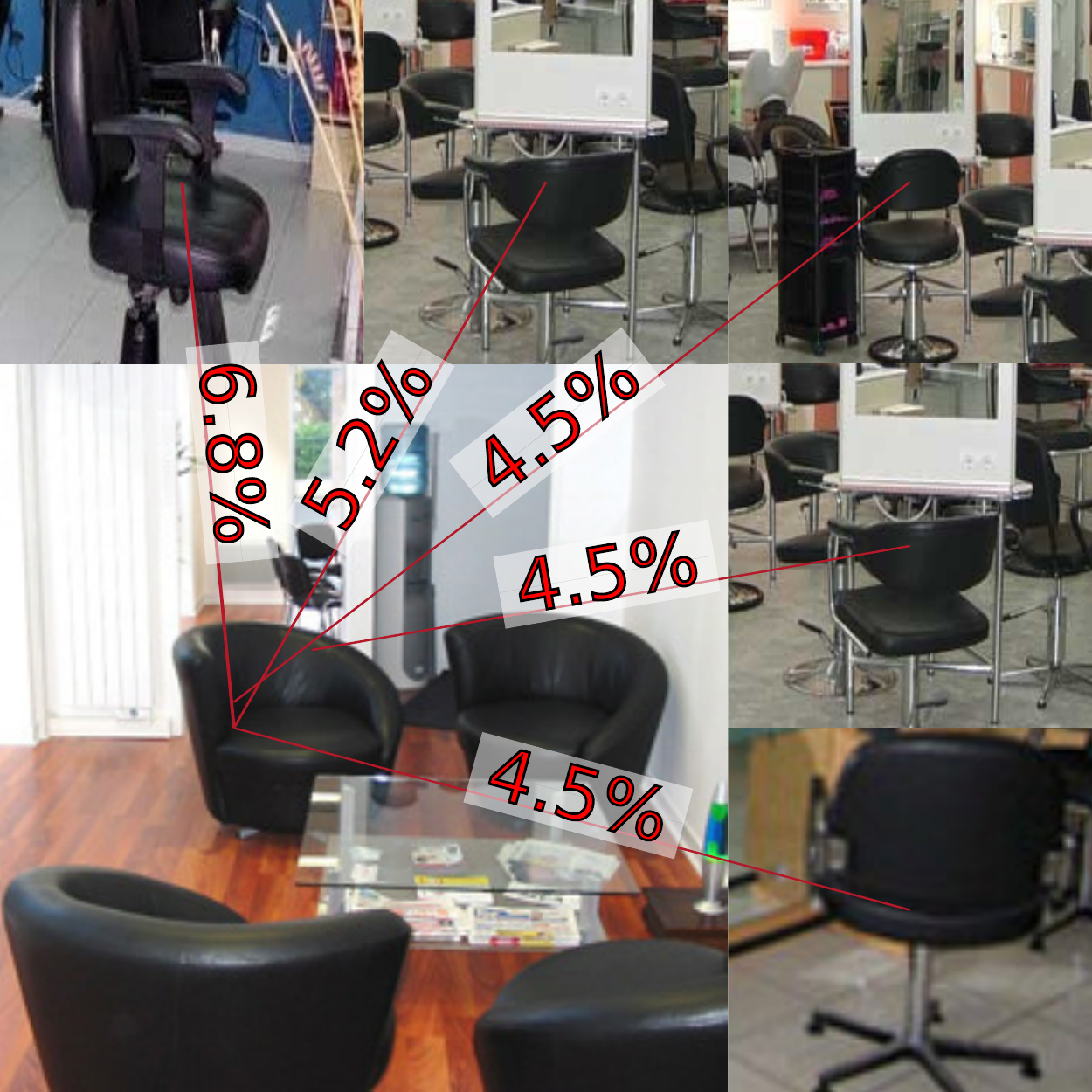}
 \includegraphics[width=0.49\linewidth]{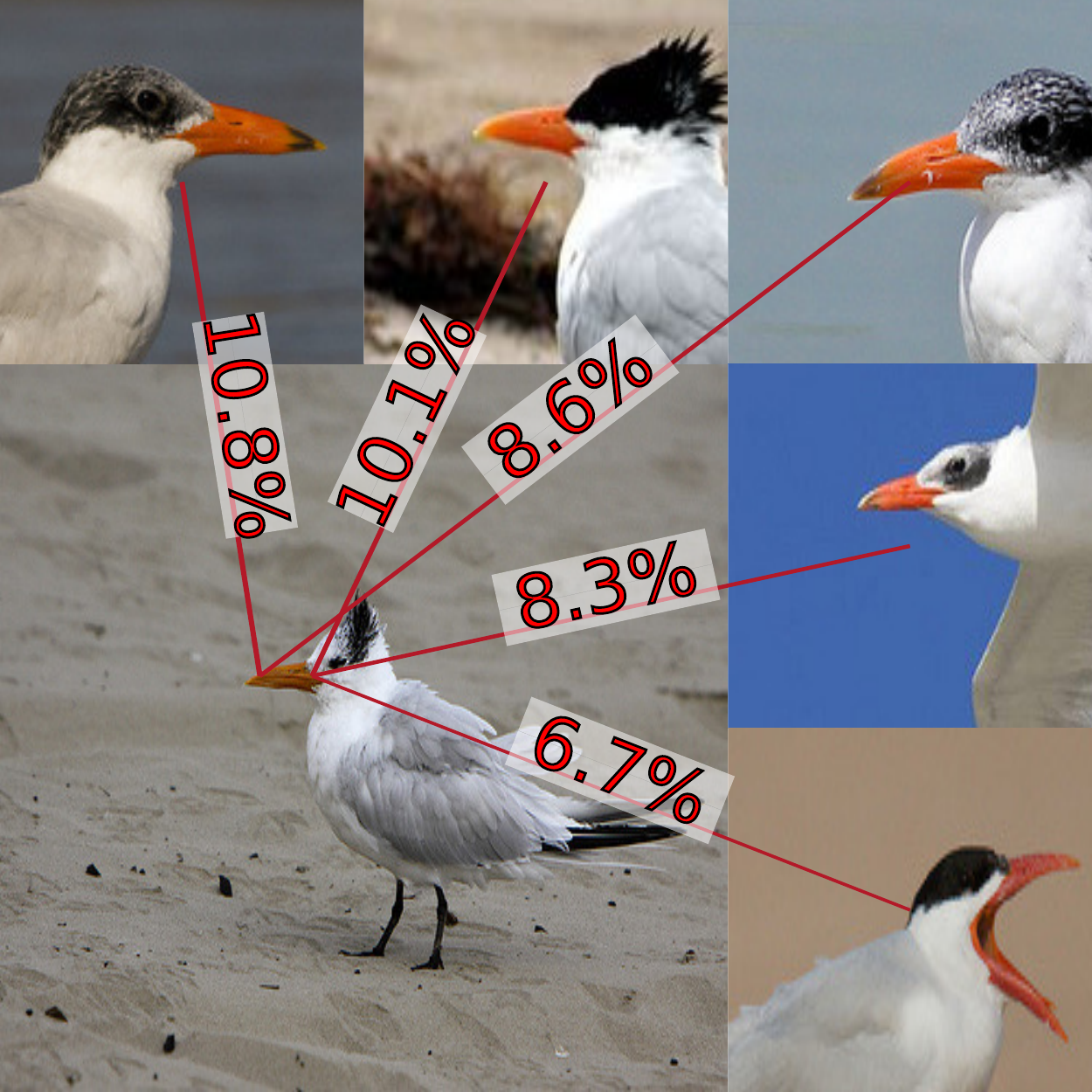}
 
 \vspace{2pt}
 \includegraphics[width=0.49\linewidth]{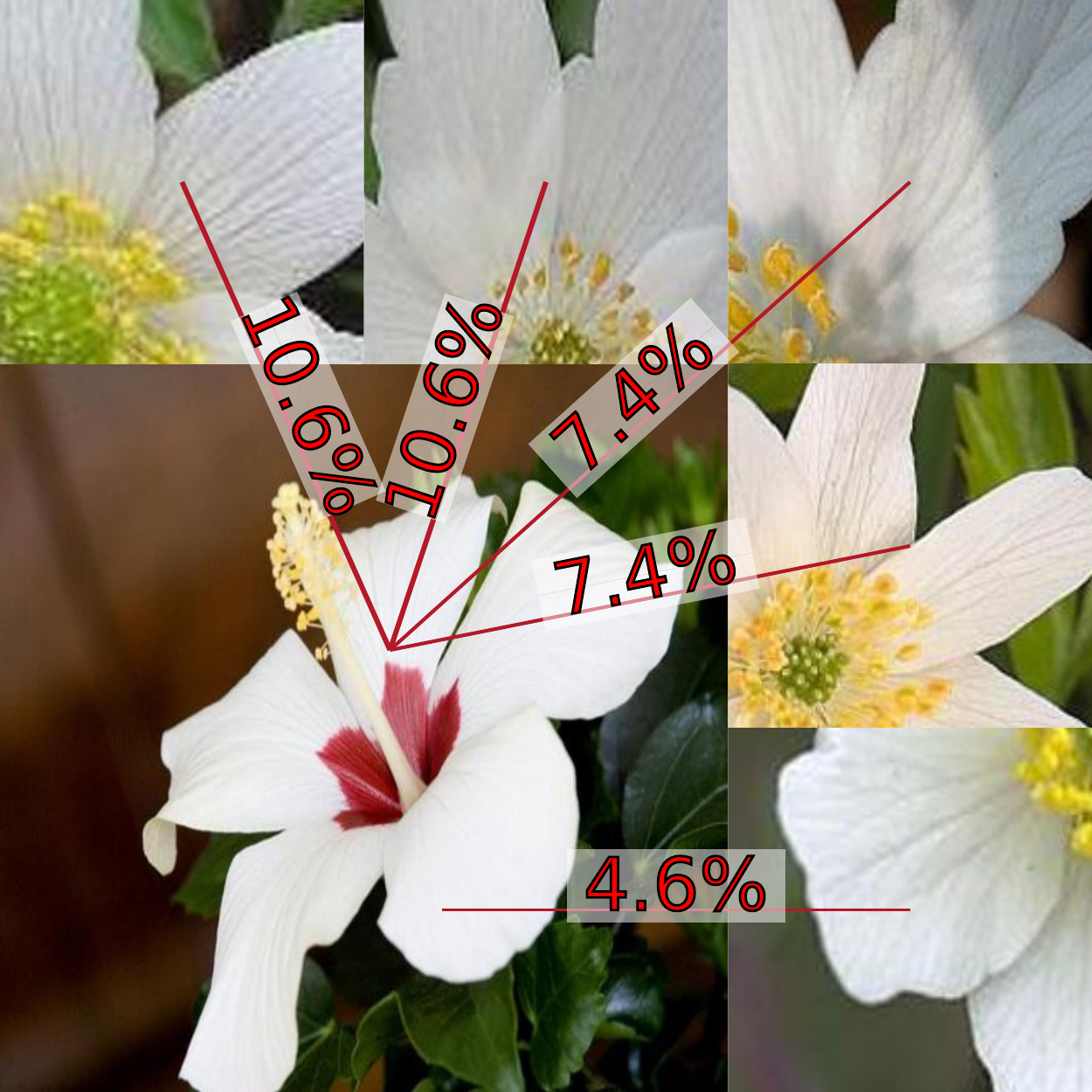}
 \includegraphics[width=0.49\linewidth]{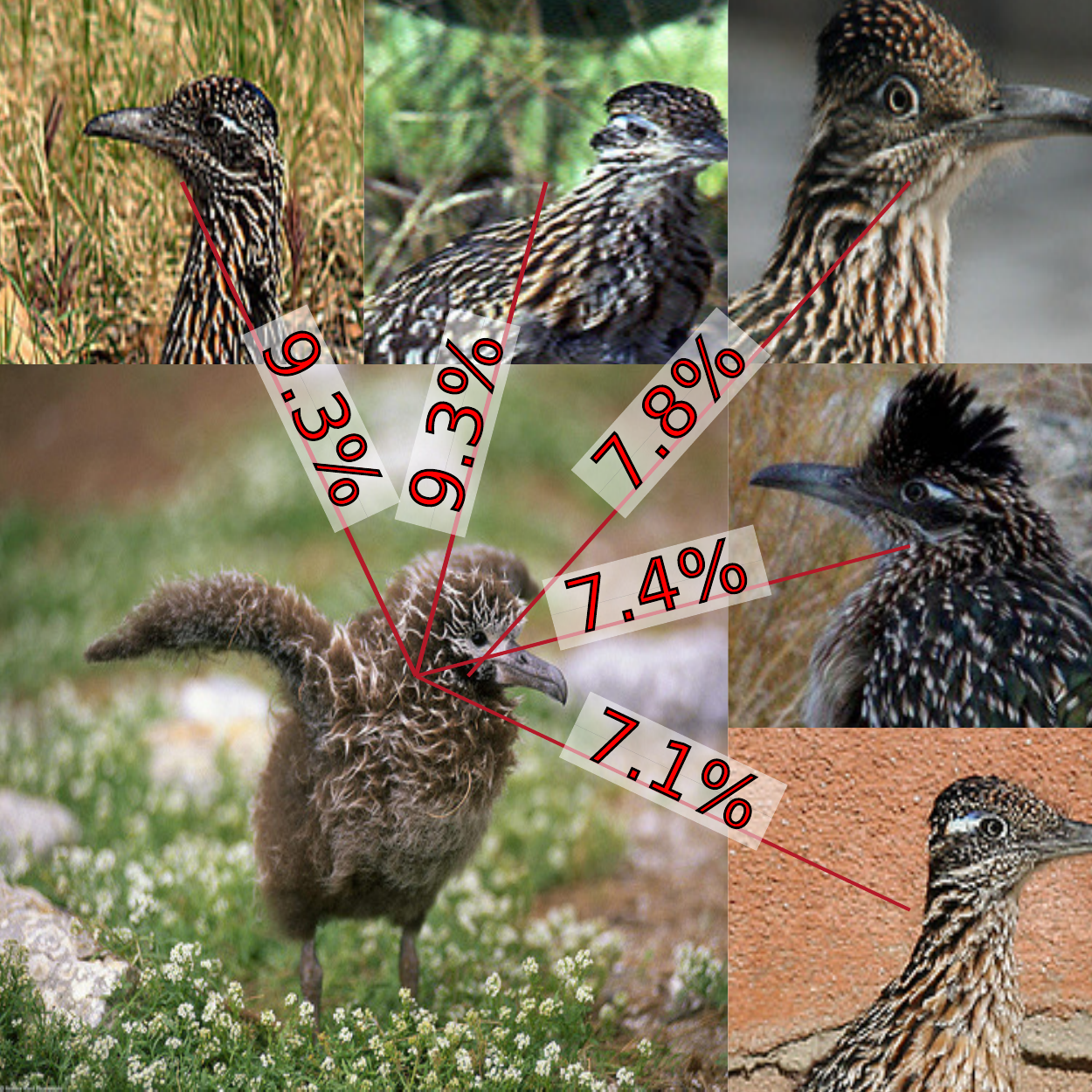}

 \caption{Visualization of the most influential image regions for the classification decision as defined in Eq.~\ref{eq:region-influence}:
    The large image in the bottom left corner is the test image and the surrounding images are crops
    of the training examples with highly relevant image regions. Percentages show the relative
    impact on the final decision. The lower four images show incorrect classifications. \label{fig:vis-classification-dec}}
\end{figure}
\figurename~\ref{fig:vis-classification-dec} depicts a classification visualization for test images from four different datasets.
In the bottom left of each block, the test image is shown.
The test image is surrounded by the five most relevant training image regions.
They are picked by first selecting the training images with the highest influence defined by the aggregated $\gamma_{k,i,j}$ over all locations $i,j$ of the test and training image.
In each training image, the highest $\gamma_{k,i,j}$ is shown using an arrow and a relative influence.
The relative influence is defined by $\gamma_{k,i,j}$ normalized by the aggregated $\gamma_{k,i,j}$ over the test and all positive training image regions, \ie images supporting the classification decision. 
Please note, that $\gamma_{k,i,j} >= 0$ for positive training samples as each element in $\inputsingle$ is greater or equal 0.
Since multiple similar triplets occur, we use non-maximum suppression and group triplets with a small normalized distance of less than $0.15$.
As can be seen, this visualization of the classification decision is intuitive and reveals the high impact of
a few small parts of the training images.

\begin{figure}
 \centering
 \includegraphics[height=0.43\linewidth]{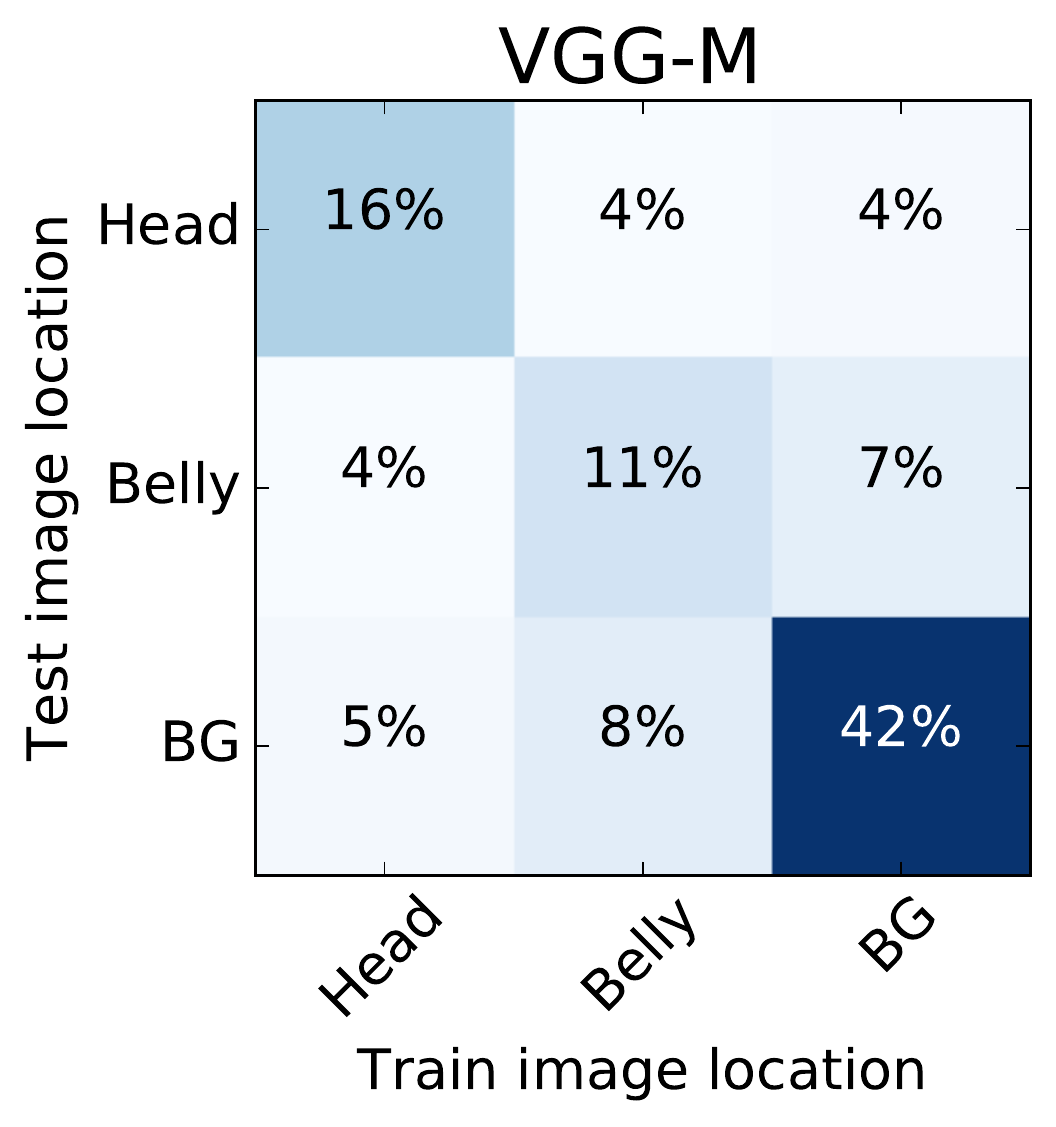}
 \hspace{10pt}
 \includegraphics[height=0.43\linewidth]{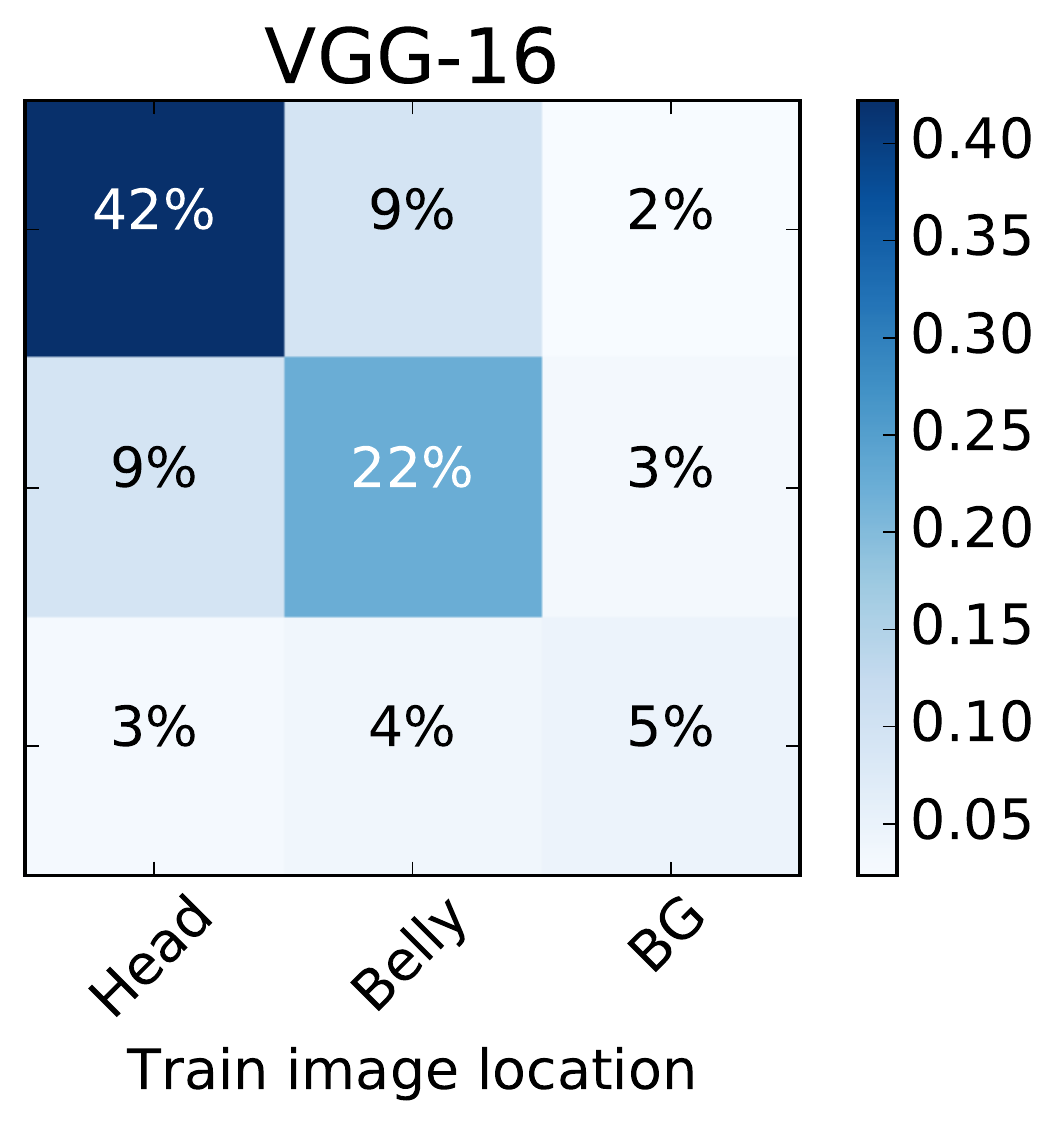}
 \caption{
 Contribution of different bird parts to the classification decision on CUB200-2011 comparing VGG-M and VGG16 without fine-tuning.
 For each semantic part in a given test image (rows), we compute the sum of inner products to another semantic part in a training image (columns).
 This statistic is normalized and averaged over all test images.
 The plots show that for VGG-M, 42\% of the classification decision can be attributed to the comparison of background elements.
 In contrast, the comparison of the bird's head is most important for VGG16.
 \label{fig:quantify-classification-dec}}
\end{figure}

\paragraph{Measuring the contribution of semantic parts}
We are also interested whether human-defined semantic parts contribute significantly to decisions.
\figurename~\ref{fig:quantify-classification-dec} shows the contribution of individual bird body parts for classification on CUB200-2011~\cite{cub200}. 
For each test image, we obtain the ten most related training image similar to before.
We divide the local feature into groups belonging to the bird's head, belly, and background and compute the sum of the squared inner products between these regions. 
As can be seen, on average, 25\% of the VGG16~\cite{simonyan14vgg16} prediction is caused by the comparison of the bird's heads.
In contrast for VGG-M~\cite{chatfield14vggm}, the background plays the most significant role with a contribution of 31\%. 
This shows that the deeper network VGG16 focuses much more on the bird instead of the background.

\paragraph{Relationship to salient regions}
We show that orderless pooling cannot just be rephrased as a correspondence kernel~\cite{gao15compactbilinear} but also as implicitly performing salient matching.
To show this, we calculate the linear kernel between the vectors $\bm{z}_k$ and $\tilde{\bm{z}}_\ell$, which induces a kernel between $\mathcal{Y}_k = \{ \inputsingle_i \}_{i=1}^{n}$ and $\mathcal{Y}_\ell = \{ \dinputsingle_j \}_{j=1}^{n}$ as follows:
\vspace{-6pt}
\begin{align}
    \notag
    \langle \poolvec_k, \tilde{\poolvec}_\ell\rangle &\propto \langle \vectorize\Bigl( \sum\limits_{i = 1}^{n} \inputsingle^{\alpha-1}_i \inputsingle_i^T \Bigr), \vectorize\Bigl( \sum\limits_{j = 1}^{n} \dinputsingle^{\alpha-1}_j \dinputsingle_j^T \Bigr) \rangle\\
    \notag
                                                                          &= \trace \Bigl( \Bigl(  \sum\limits_{i = 1}^{n} \inputsingle_i^{\alpha-1} \inputsingle_i^T \Bigr)^T \Bigl(  \sum\limits_{j = 1}^{n} \dinputsingle_j^{\alpha-1} \dinputsingle_j^T \Bigr) \Bigr)\\
    &= \sum\limits_{i,j} \langle \inputsingle_i, \dinputsingle_j \rangle \cdot \langle \inputsingle_i^{\alpha-1}, \dinputsingle_j^{\alpha-1} \rangle, \label{eq:pairwise-matching}
\end{align}
where we ignored normalizing with respect to $n$ for brevity. 
Please note that this derivation also reveals that the difference between bilinear and average pooling is only the quadratic transformation of the scalar product between two feature vectors $\inputsingle_i$ and $\dinputsingle_j$. 

We can now show a further direct relation to a simple matching of local features in two images by rewriting the scalar products as:
\vspace{-4pt}
\begin{align}
\label{eq:alphalin}
    \langle \inputsingle_i, \dinputsingle_j \rangle &\propto \sum\limits_{i,j} \left( \|\inputsingle_i\|^2 + \|\inputsingle_j\|^2 - \| \inputsingle_i - \inputsingle_j \|^2 \right)\enspace,
\end{align}
where the Euclidean distance between two features appears.
\begin{figure}
  \centering 
   \includegraphics[width=0.85\linewidth]{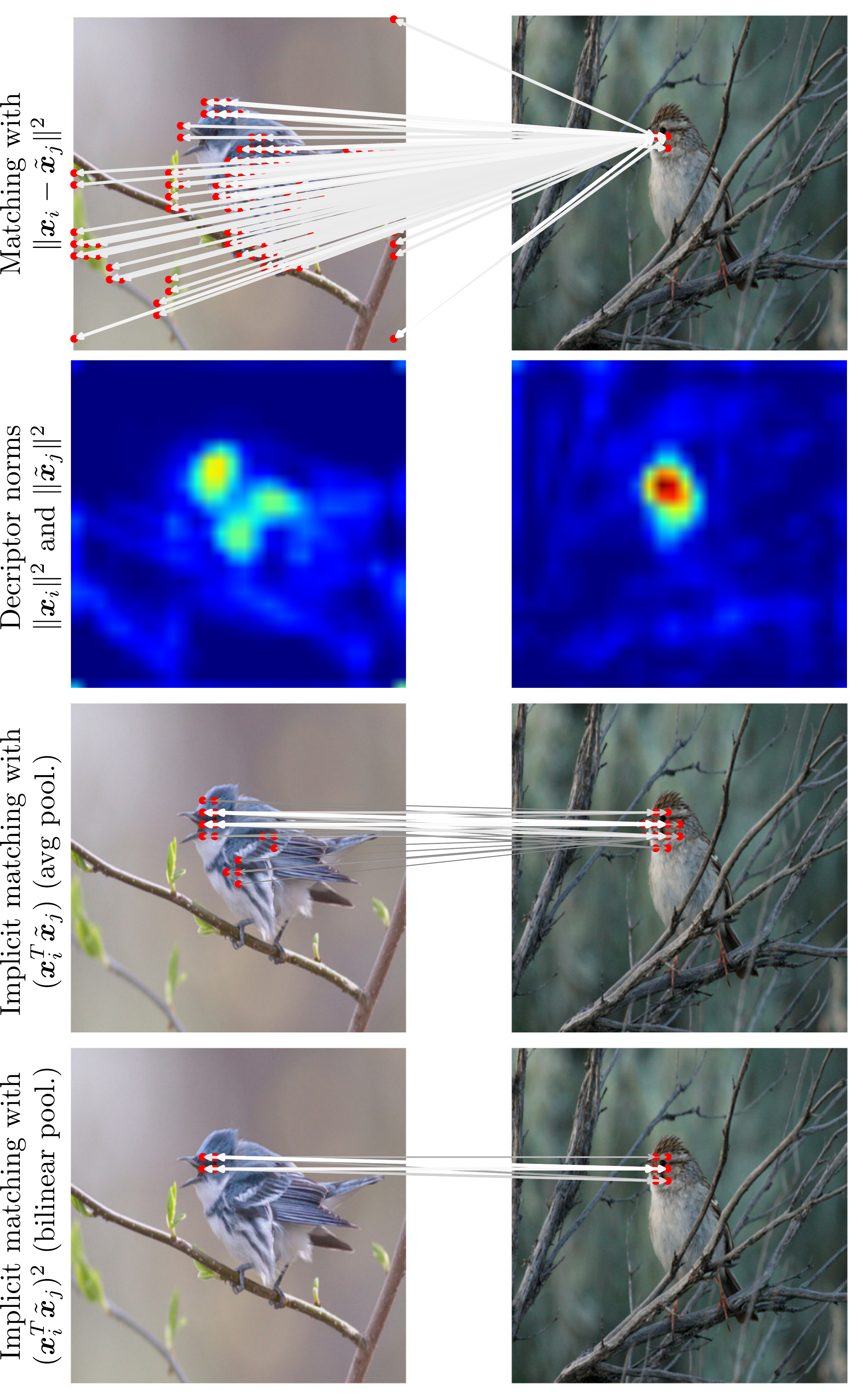}

  \caption{Visualization of the pairs of most similar local features using L2-distance.
  The thicker and whiter the line, the more similar are the features.
  In the first row, feature similarity was defined by the lowest L2-distance between local features.
  The second row shows the magnitude of all local features normalized to the highest norm in both feature maps.
  The third and fourth row show the implicit matchings using the inner product and the squared inner product as similarity measure, 
  as it is used in average and bilinear pooling. We only show matchings larger than 50\% of the maximum matching in this case.
  VGG16 with increased input size of $448\times 448$ and the output of \texttt{conv5\_3} after activation and before pooling was used. 
  The local features have a spatial resolution of $28\times 28$.
  \label{fig:l2norm-matching}
  }
\end{figure}
The kernel output is therefore high if the feature vectors are highly similar (small Euclidean distance) especially
for pairs $(i,j)$ characterized by individual high Euclidean norms. In \figurename~\ref{fig:l2norm-matching}, we visualize the Euclidean norms of the
feature vectors in \texttt{conv5\_3} extracted with VGG16. As can be observed when comparing the plot for the norm and the matching, areas with a high magnitude of the features also correspond to salient
regions. This is indeed reasonable since the ``matching cost'' in Eq.~\eqref{eq:pairwise-matching} should focus on the relevant object itself
and not on background elements.

\paragraph{Focusing pairwise similarities by increasing $\bm{\alpha}$}
Similar to \cite{lin15bilinear}, we apply pooling directly after the ReLU activation following
the last convolutional layer.
Therefore, all scalar products between these features are positive.
Hence, summands with a high scalar product are emphasized dramatically for large values of $\alpha$ and in particular also for bilinear pooling.
Increasing $\alpha$ therefore leads to kernel values likely based on only a few relevant pairs $(i,j)$.
This fact is illustrated in the last two rows of \figurename~\ref{fig:l2norm-matching}, where we only showed the pairs
with an inner product larger $50\%$ of the highest one for both average and bilinear pooling.

\section{Experimental Analysis \label{sec:experiments}}

In our experiments, we focus on analyzing the difference between average and bilinear pooling for image recognition.
We make use of our novel $\alpha$-pooling, which was presented in Section~\ref{sec:generalizing}.
First, we show that it achieves state-of-the-art results in both generic and fine-grained image recognition.
In these experiments, $\alpha$ is learned from data and hence the pooling strategy is determined by the data.
Second, based on deep neural nets learned on both kinds of datasets, we analyze distinguishing properties using the visualization techniques presented in Section~\ref{sec:pairwise-matching}.
In detail, we discuss the relationship of the parameter $\alpha$ with dataset granularity, classification decisions, and implicit pose normalization.
Hence we manually set the value of $\alpha$ in this second part.

\begin{figure}
\begin{minipage}[t]{0.48\linewidth}
  \includegraphics[width=\linewidth]{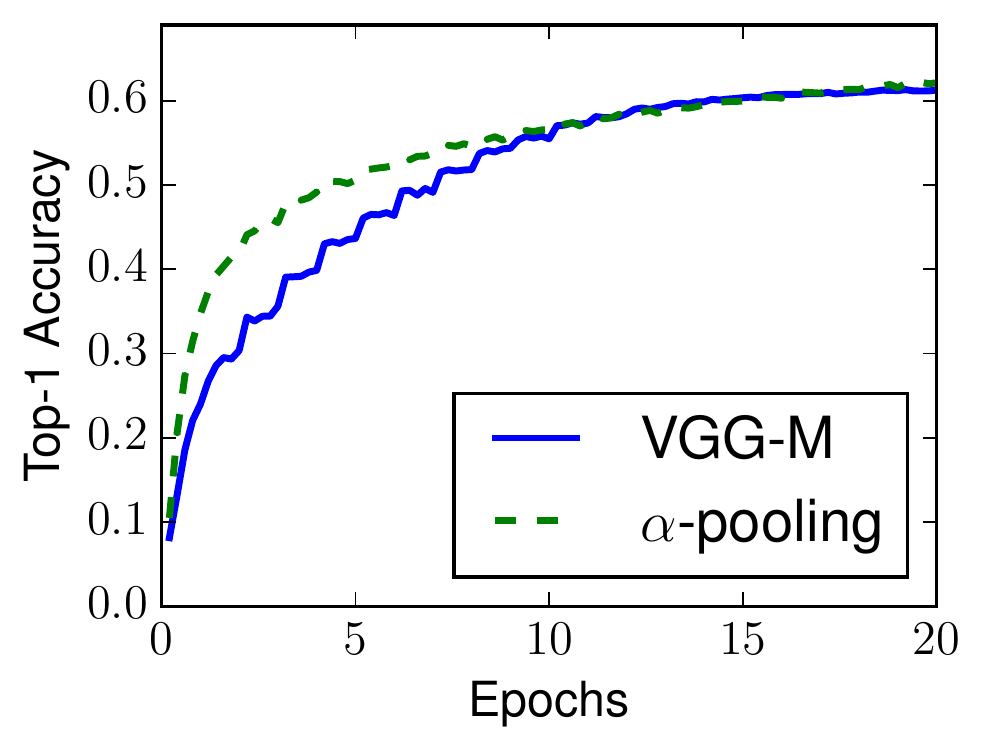}
\end{minipage}%
    \hfill%
\begin{minipage}[t]{0.48\linewidth}
    \vspace{-2.8cm}
    \resizebox{\linewidth}{!}{
      \begin{tabular}{lcc}
      \toprule
      Approach & Top-1 accuracy \\ 
	& (Single crop) \\ 
      \midrule
      \vspace{0.5em}VGG-M~\cite{chatfield14vggm}&63.1\% \\
      \vspace{0.5em}Bilinear~\cite{lin16texture}& 63.4\% \\
      \textit{Our work:}\\
      VGG-M w/ $\alpha$-pool & \textbf{64.2\%} \\
      \bottomrule
      \end{tabular}
    }
\end{minipage} 
\caption{Accuracy on ImageNet using $\alpha$-pooling for VGG-M. 
The left plot shows validation accuracy over the first twenty trained epochs. 
VGG-M denotes the original architecture with batch normalization added and two fully-connected layers before the classifier. 
$\alpha$-pooling is the novel generalized pooling technique, which replaces these two fully-connected layers.
With $\alpha$-pooling, the network converges faster and achieves a higher validation accuracy.
$\alpha$ is learned from data.
  \label{fig:imagenet-results}
}
\end{figure}

\paragraph{Accuracy of $\bm{\alpha}$-pooling}
We evaluated both training from scratch and fine-tuning using a network pre-trained on ImageNet Large Scale Visual Recognition 2012 challenge dataset~\cite{russakovsky15imagenet}.
For training from scratch, we used the VGG-M~\cite{chatfield14vggm} architecture and replaced the last pooling and all fully-connected layers with $\alpha$-pooling.
Batch normalization~\cite{ioffe15batchnorm} was used after all convolutions as well as after the $\alpha$-pooling.
In addition, we used dropout with a probability of $p=0.5$ to reduce overfitting and increase generalization ability.
Compact bilinear encoding~\cite{gao15compactbilinear} was used to compute a compact representation of the aggregated outer products with a dimensionality of 8096.
The learning rate started with 0.025 and followed an exponential decay after every epoch. 
The batch size was set to 256.
The results on ILSVRC 2012 (1000 classes, 1.2 million images) are shown in \figurename~\ref{fig:imagenet-results}.
We plot both the validation accuracy during the first twenty epochs of the training as well as the final top-1 single crop accuracy.
As can be seen, the network converges faster at only small additional computation cost.
In addition, it reaches a higher final accuracy compared to the original VGG-M with batch normalization.

For fine-tuning, we use a VGG-M~\cite{chatfield14vggm} and VGG16~\cite{simonyan14vgg16} pre-trained on ILSVRC 2012~\cite{russakovsky15imagenet} and replace all layers after the last convolutional layer with the novel $\alpha$-pooling encoding.
We follow the authors of \cite{lin15bilinear,gao15compactbilinear} and add a signed square root as well as $L_2$-normalization layer before the last linear layer.
Pooling is done across two scales with the smaller side of the input image being 224 and 560 pixels long.
Two-step fine-tuning~\cite{branson14posenorm} is used, where the last linear layer is trained first with $0.01$ times the usual weight decay and the whole network is trained afterwards with the usual weight decay of $0.0005$.
The learning rate is fixed at $0.001$ with a batch size of eight.
$\alpha$ is learned from data.

The results for CUB200-2011 birds~\cite{cub200}, FGVC-Aircraft~\cite{aircraft} and Stanford 40 actions~\cite{40actions} can be seen in \tablename~\ref{tab:bilinear-over-scale}.
    \begin{table}
    \caption{Accuracy on several datasets with $\alpha$-pooling using the multi-scale variant. 
      No ground-truth part or bounding box annotations were used.
      $\alpha$ is learned from data. 
      \label{tab:bilinear-over-scale}
    }
	  \centering
	  \resizebox{1.0\linewidth}{!}{
	    \begin{tabular}{lccc}
	    \toprule
	    Dataset & CUB200-2011 & Aircraft & 40 actions\\ 
	    classes / images& \small{200 / 12k} & \small{89 / 10k}& \small{40 / 9.5k}\\
	    \midrule 
	    Previous & 81.0\%~\cite{simon15nac} & 72.5\%~\cite{chai13symbiotic} & 72.0\%~\cite{zhou16deep}\\
	     & 82.0\%~\cite{krause15finegrained} & 78.0\%~\cite{rosenfeld16iterative} &80.9\%~\cite{cai16prop} \\
	    \vspace{5pt} & 84.5\%~\cite{zhang16picking} & 80.7\%~\cite{gosselin14fisher} & 81.7\%~\cite{rosenfeld16iterative} \\
	    Special case: bilinear~\cite{lin15bilinear} & 84.1\% & 84.1\% & - \\
	    Learned strategy (Ours)& \textbf{85.3\%} & \textbf{85.5\%}& \textbf{86.0\%}\\ 
	    \bottomrule
	    \end{tabular}
	  }
    \end{table}
We achieve higher top-1 accuracy for all datasets compared to previous work. 
For fine-grained datasets like birds and aircraft, we slightly improve the results of \cite{lin15bilinear,gao15compactbilinear}, which is due to $\alpha=2$ being close to the learned $\alpha$ for this dataset.
Our approach also shows a high accuracy on datasets besides traditional fine-grained tasks as shown by the Stanford 40 actions dataset, where we achieve 83.5\% accuracy compared to the 81.7\% reported in \cite{rosenfeld16iterative}.

\paragraph{Ranking dataset granularity wrt.~$\bm{\alpha}$}
As already mentioned, the one main purpose of these experiments is to analyze the differences between generic and fine-grained image classification tasks.
In particular, we are interested in why average pooling lacks accuracy in fine-grained tasks while bilinear can reach state-of-the-art. 

\begin{figure}
 \centering
\begin{minipage}[t]{0.48\linewidth}
  \includegraphics[width=\linewidth]{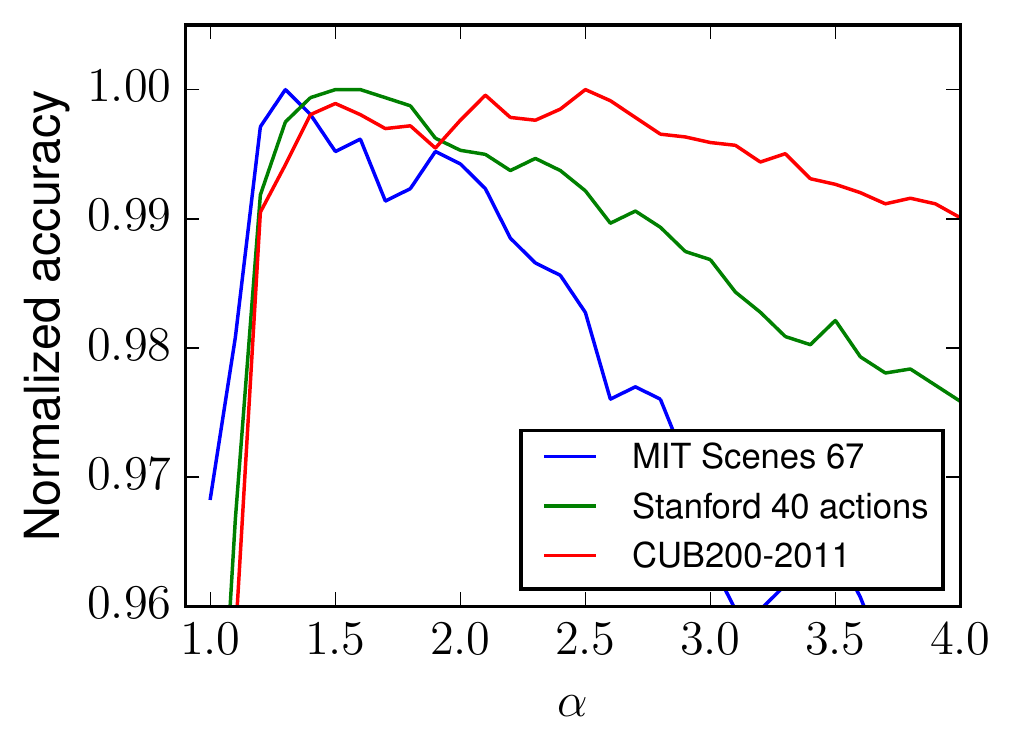}
\end{minipage}%
    \hfill%
\begin{minipage}[t]{0.48\linewidth}
  \centering 
  \vspace{-3cm}
  \includegraphics[width=\linewidth]{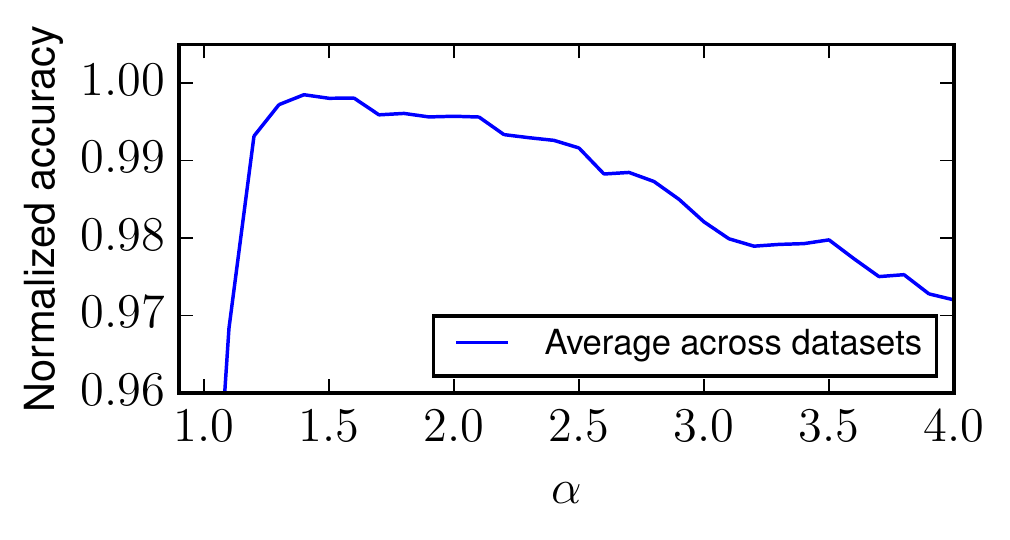}
   \includegraphics[width=\linewidth]{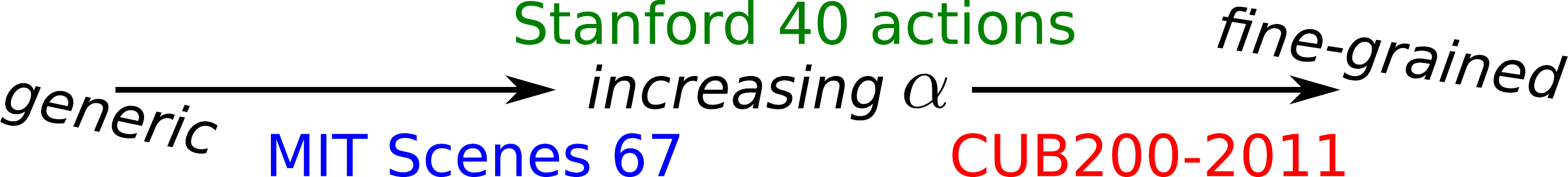}
\end{minipage} 
 \caption{Influence of $\alpha$ using VGG16 without fine-tuning. 
 $\alpha=1$ corresponds to average pooling and $\alpha=2$ to bilinear pooling.
 $\alpha$ is set manually in this experiment.
 \label{fig:alpha_pooling}
 }
\end{figure}

The presented $\alpha$-pooling allows for a smooth transition between average and bilinear pooling. 
In this paragraph, we analyze the relationship of the parameter $\alpha$ and the granularity of the dataset. 
The results using VGG16 without fine-tuning can be seen in \figurename~\ref{fig:alpha_pooling}.
The input resolution of both networks was increased to $448\times 448$ to achieve state-of-the-art results without fine-tuning. 
For each dataset and value of $\alpha$, we train with a multinomial logistic loss.
The accuracy is normalized to 1 for each dataset for easier comparison of different datasets.
As can be seen, each dataset seems to require a different type of pooling.
If the datasets are ordered by the value of $\alpha$, which gives the highest validation accuracy, the order is as follows: MIT Scenes 67~\cite{quattoni09mit67}, 40 actions~\cite{40actions}, and CUB200-2011~\cite{cub200} with $\alpha=1.3$, $1.5$, and $2.5$, respectively.
This seems to suggest that the more we move from generic to fine-grained classification, the higher is the value of $\alpha$.
In addition, larger values of alpha are still good for fine-grained while accuracy drops quickly for generic tasks. 
Hence focusing the classification on few correspondences seems a good strategy for fine-grained while it lowers accuracy on generic tasks.
VGG-M shows a similar trend. %

\begin{figure}
 \begin{minipage}[t]{0.48\linewidth}
 \centering
 $\alpha=1$ 
 
 \includegraphics[width=0.9\linewidth]{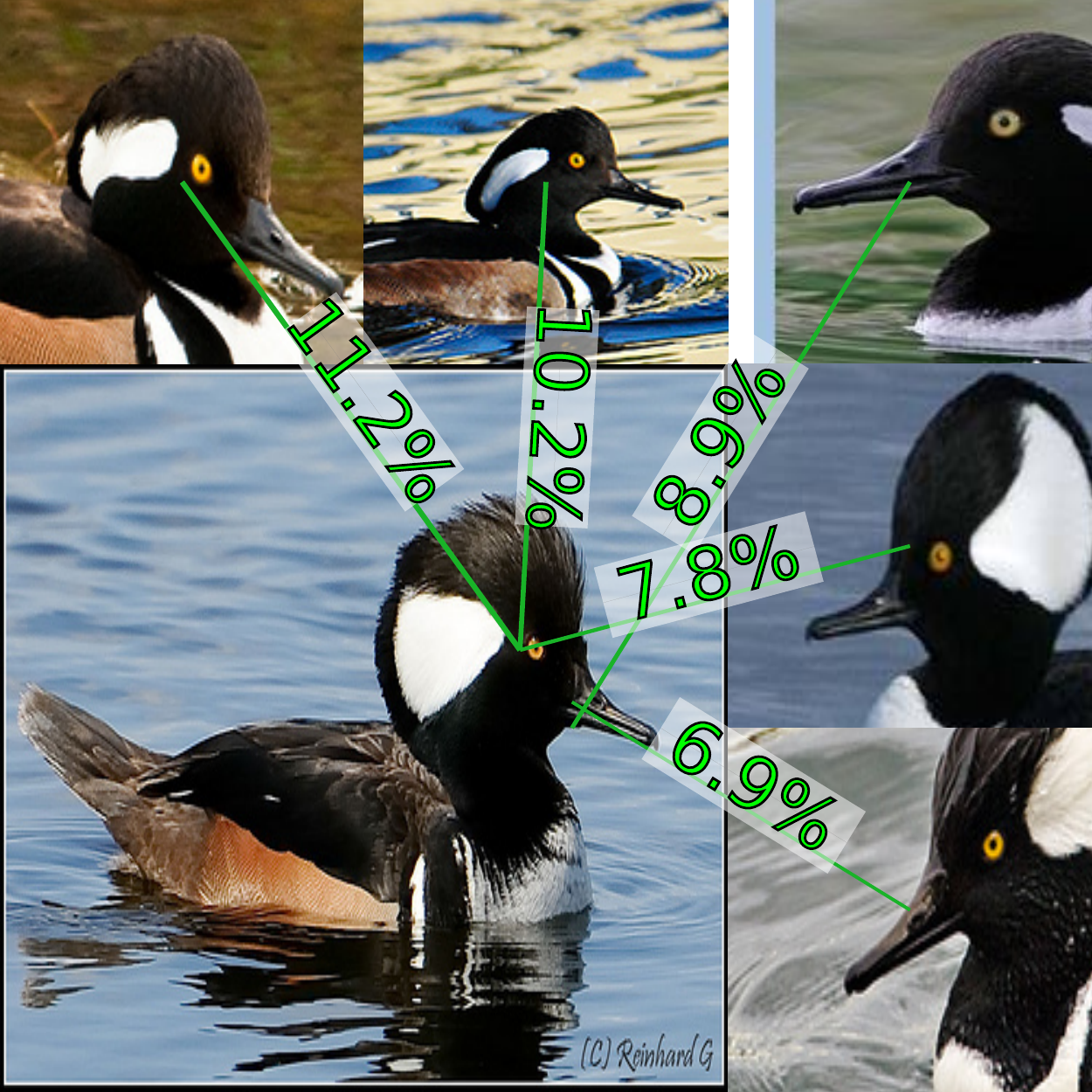}
 
 \includegraphics[width=0.9\linewidth]{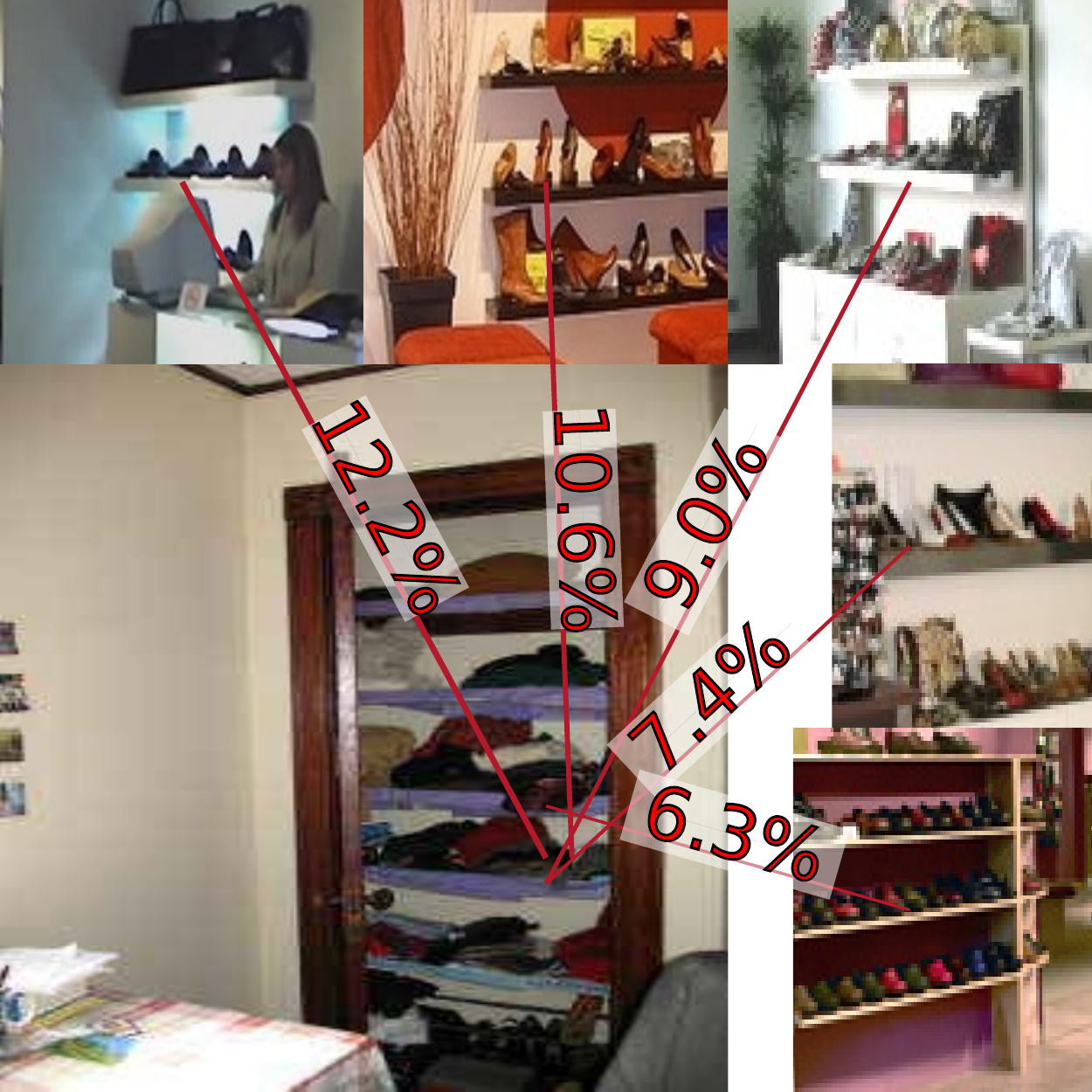}
 
\end{minipage}%
    \hfill%
\begin{minipage}[t]{0.48\linewidth}
 \centering
 $\alpha=3$
 
 \includegraphics[width=0.9\linewidth]{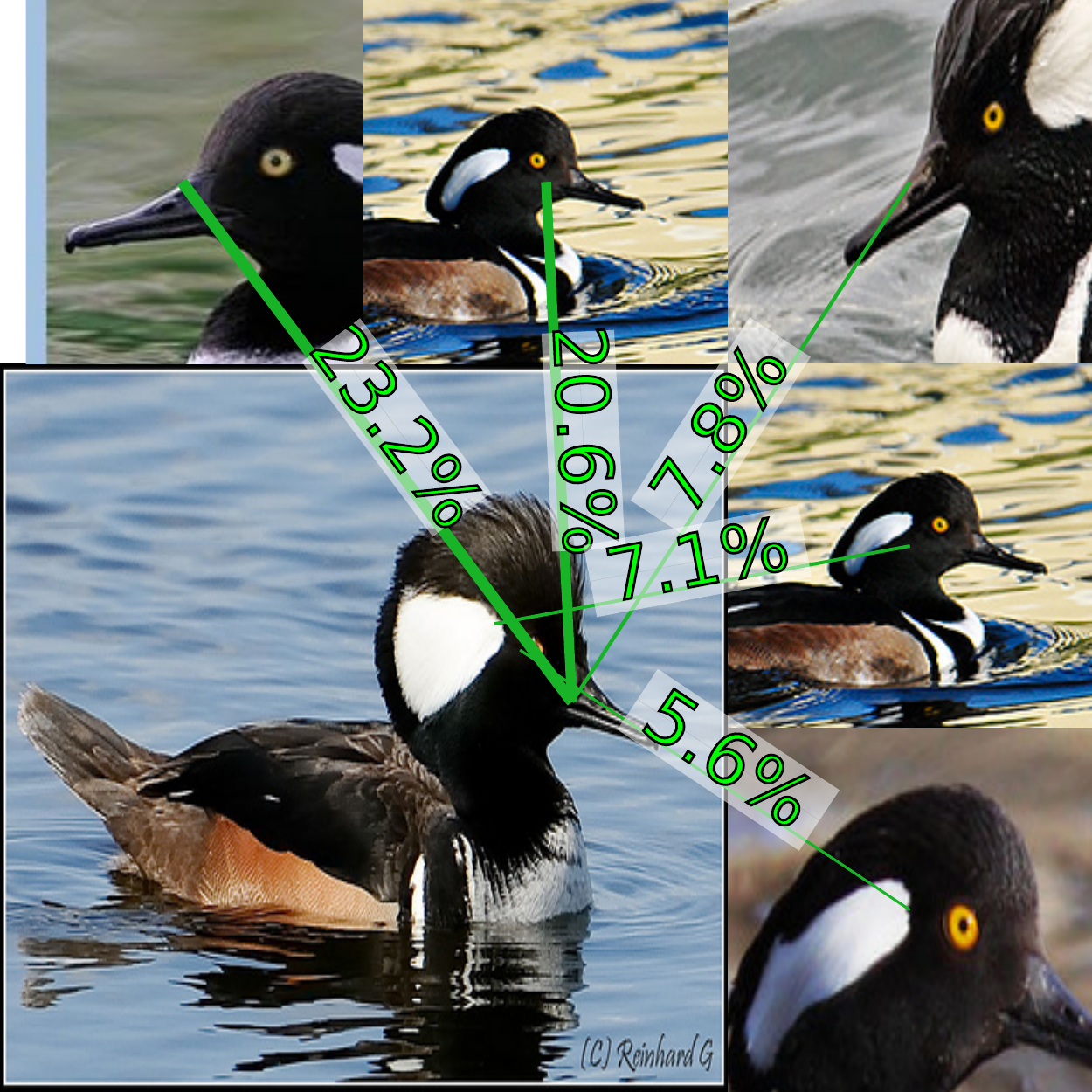}
 
 \includegraphics[width=0.9\linewidth]{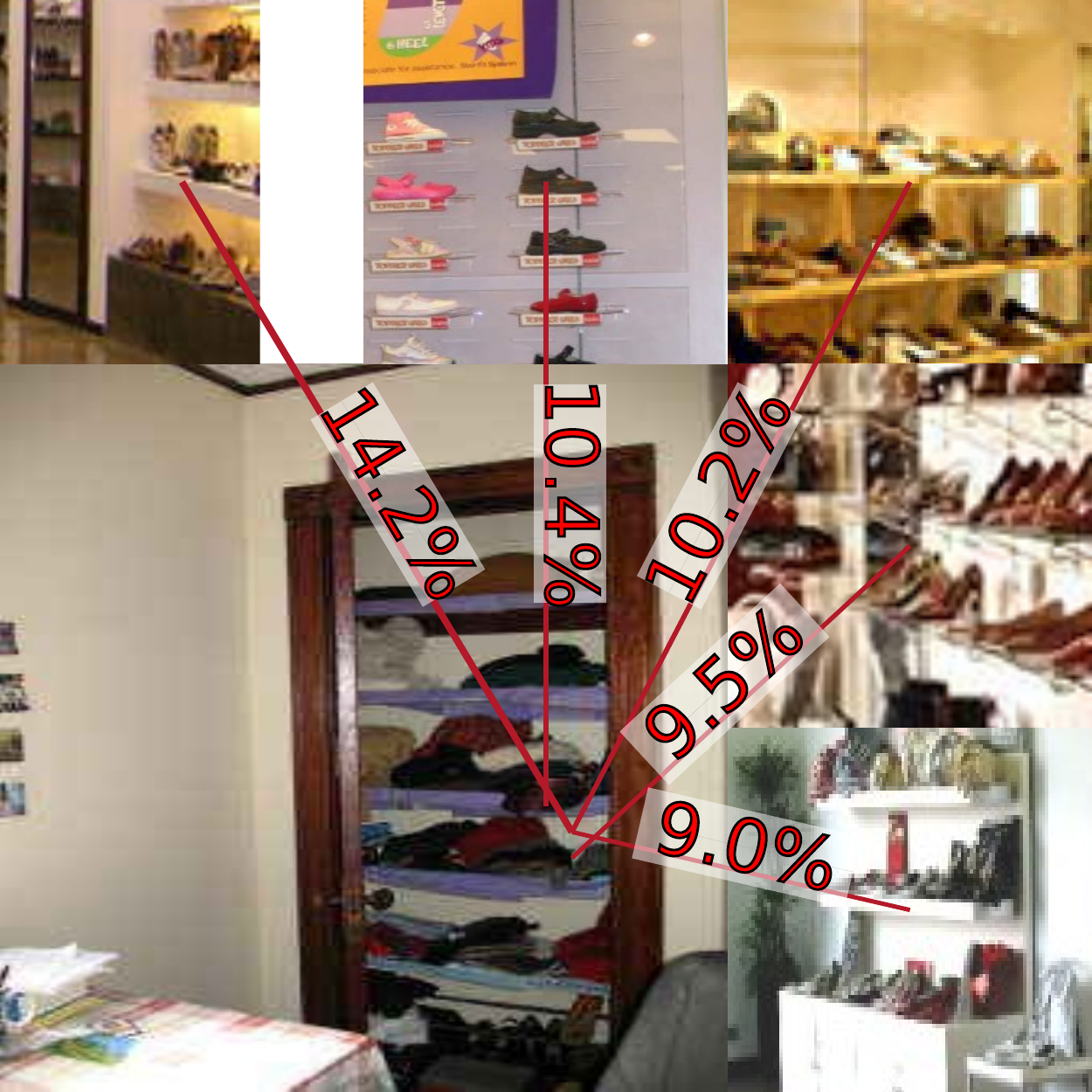}
 
\end{minipage} 
\vspace{5pt}
 \caption{Influence of $\alpha$ on the classification decisions. 
 As in \figurename~\ref{fig:vis-classification-dec}, we visualize the most relevant corresponding image regions for the classification decision.
 A larger $\alpha$ increases the importance of the most related regions in the training images.
 Hence the decision is based on few important regions.
 $\alpha$ is set manually in this experiment.
 \label{fig:alpha-vs-classification-dec}}
\end{figure}

\paragraph{Classification visualization versus $\bm{\alpha}$}
Section~\ref{sec:pairwise-matching} presented a novel way to visualize classification decisions for feature representations based on $\alpha$-pooling.
We are now interested in the change of classification decision reasoning with respect to $\alpha$.
\figurename~\ref{fig:alpha-vs-classification-dec} shows the classification visualization for two sample test images from CUB200-2011 and MIT scenes 67.
For each test image, we show the visualization for $\alpha=1$ and $\alpha=3$. 
While $\alpha=1$ causes a fairly equal contribution of multiple training image regions to the decision, $\alpha=3$ pushes the importance of the first few images.
For example, the contribution of the most relevant training image region grows from 11.2\% to 23.2\% in case of the bird image.
A statistical analysis for all test images and datasets can be found in the supplementary material.

\paragraph{Relevance of semantic parts versus $\bm{\alpha}$}
A second way to analyze the pairwise matching induced by $\alpha$-pooling is to quantify the matchings between ground-truth parts.
Similar to the previous paragraph, we are especially interested in why average pooling is not well suited for fine-grained classification on CUB200-2011.

We evaluated on CUB200-2011 using the ground-truth part annotations.
A ground-truth segmentation of the bird's head and belly was generated based on these annotations and used to assign feature locations in \texttt{conv5\_3} of VGG16 to different body parts. 
Afterwards, for each test image, the kernel between all features of the bird's head in the test and a training image is aggregated.
This is done for all pairs of regions and for the 10 most relevant training images.
The obtained statistics are then averaged over all test images.

\begin{figure}
 \centering
 
 \includegraphics[height=0.34\linewidth]{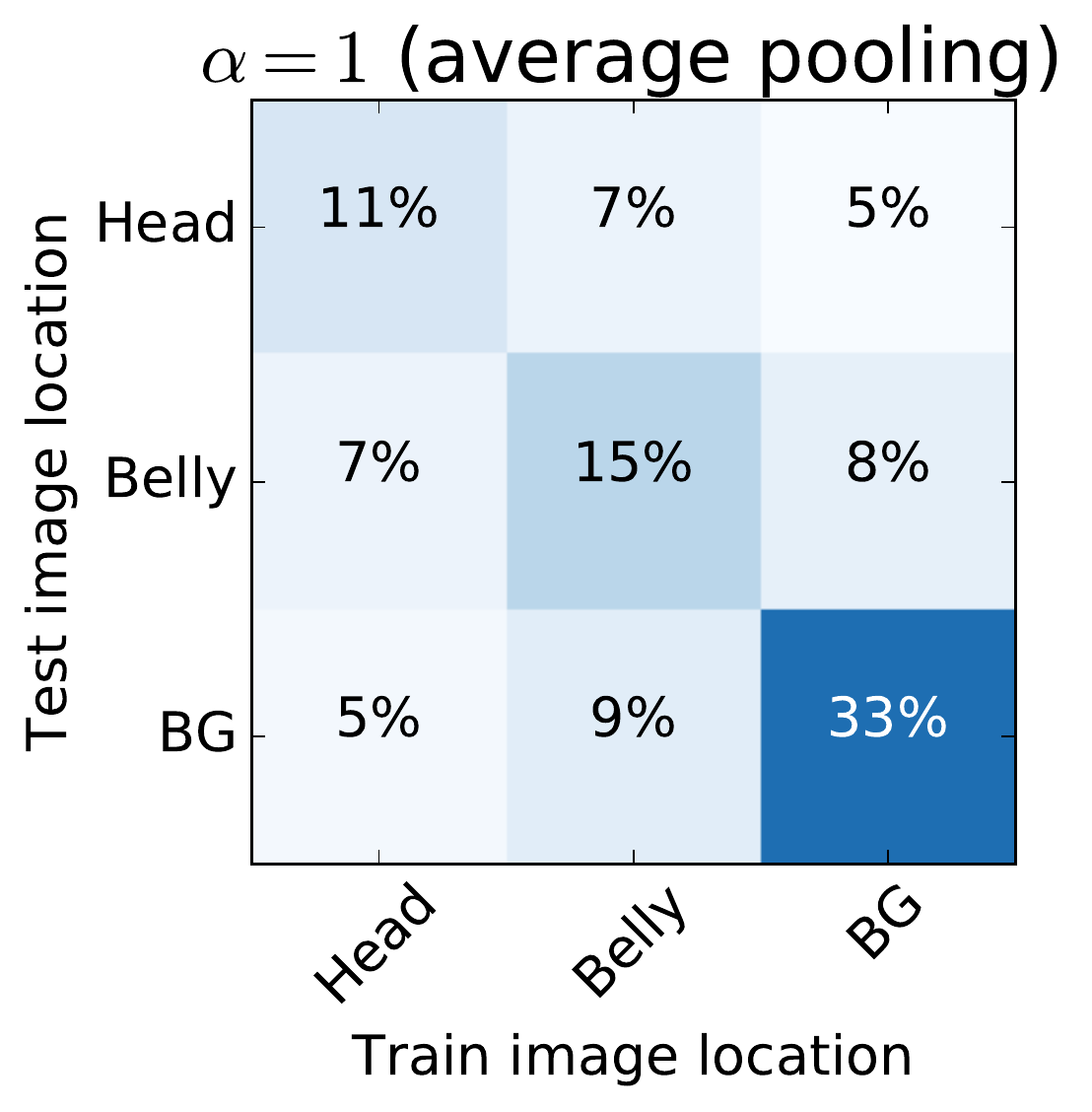}
 \hspace{0pt}
 \includegraphics[height=0.34\linewidth]{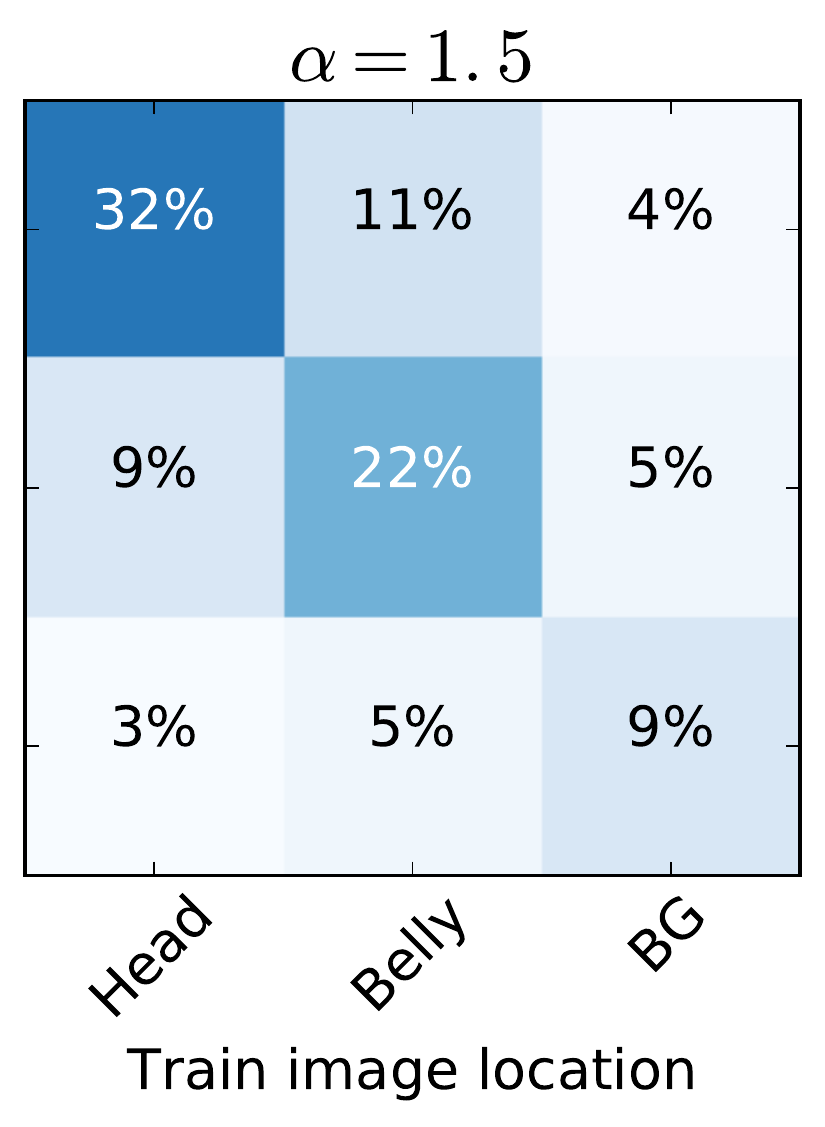}
 \hspace{0pt}
 \includegraphics[height=0.34\linewidth]{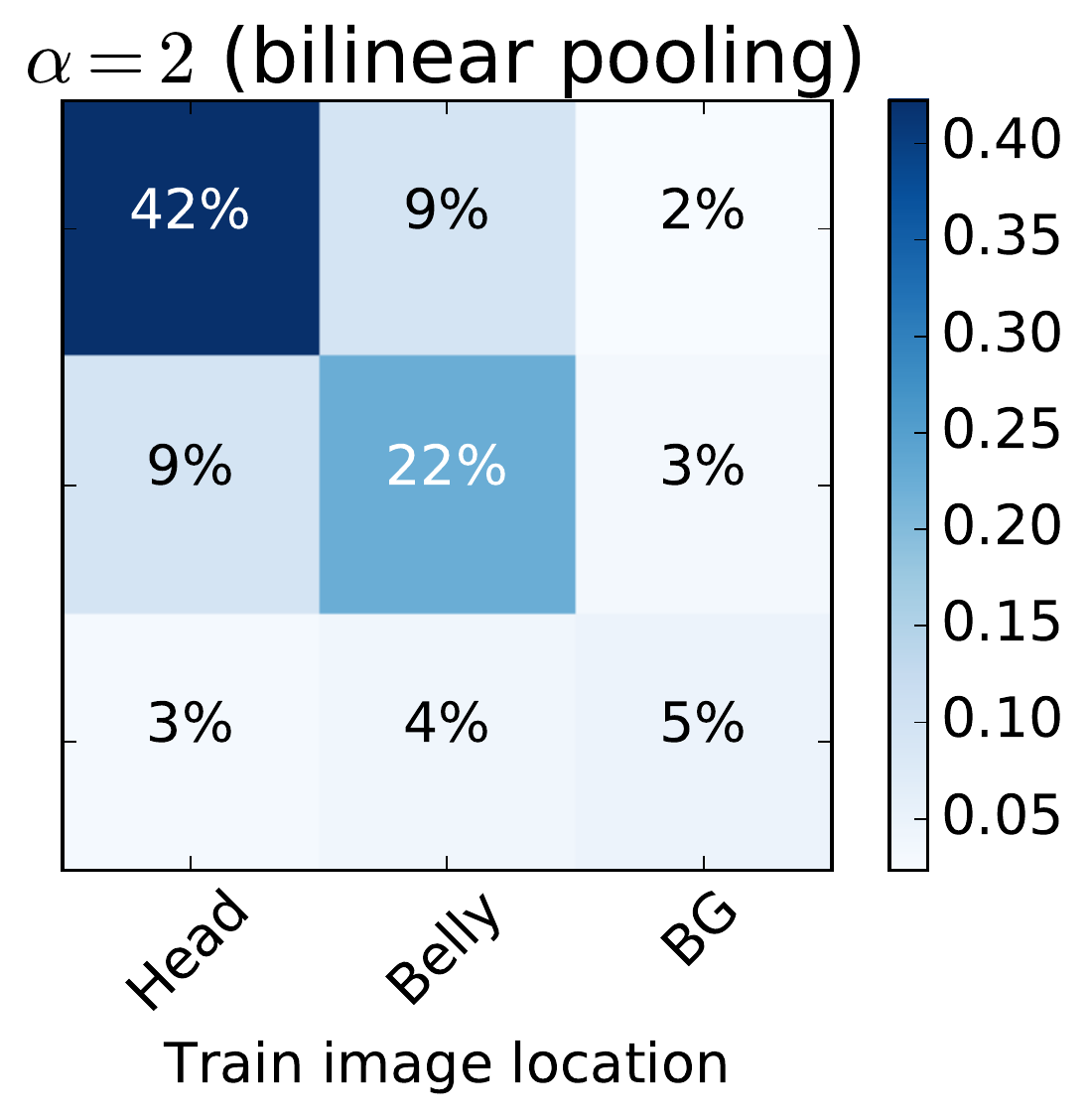}

 \vspace{5pt}
 \includegraphics[width=0.8\linewidth]{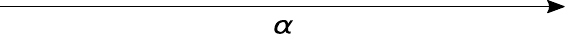}
 \caption{
 Influence of $\alpha$ on the contribution of different bird body parts to the classification decision on CUB200-2011.
 The higher the value of $\alpha$, the higher is the influence of the actual bird body parts to the classification decision.
 $\alpha$ is set manually in this experiment.
 \label{fig:quantify-classification-vs-alpha}}
\end{figure}

First, we analyze the influence of $\alpha$ on the contribution of different body parts.
\figurename~\ref{fig:quantify-classification-vs-alpha} shows the results for VGG16 without fine-tuning and $\alpha \in \{1,1.5,2\}$.
The plot shows, how a larger value of $\alpha$ focuses the classification decision more on the body parts of the bird. 
The contribution of the bird's head to the classification decision shifts from 9\% ($\alpha=1$) to 42\% ($\alpha=2$).
This observation matches our previous interpretation that larger values for $\alpha$ focus the classification decision on fewer discriminative pairs of local features. 

\begin{figure}
 \centering
 \includegraphics[height=0.43\linewidth]{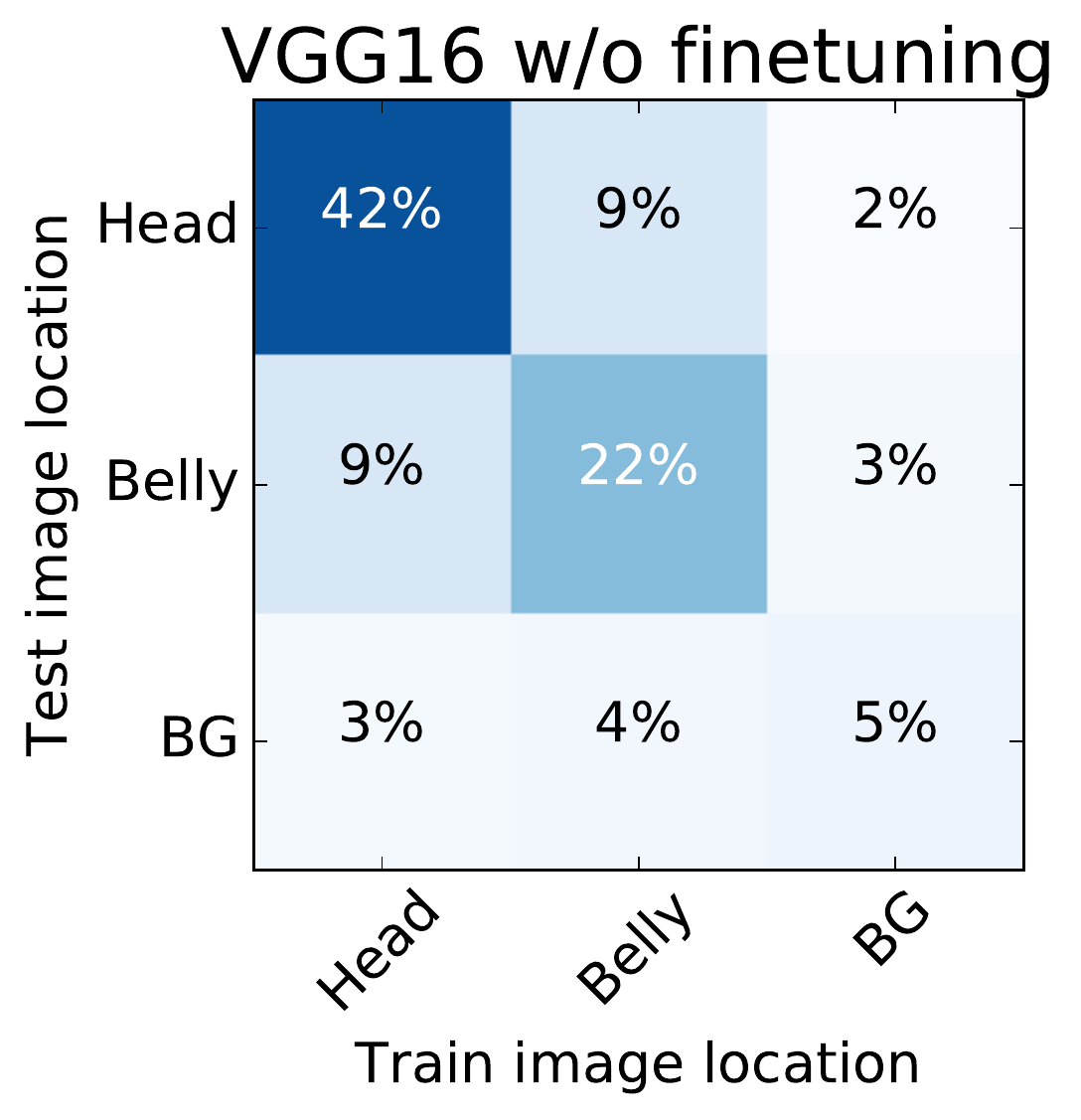}
 \hspace{10pt}
 \includegraphics[height=0.43\linewidth]{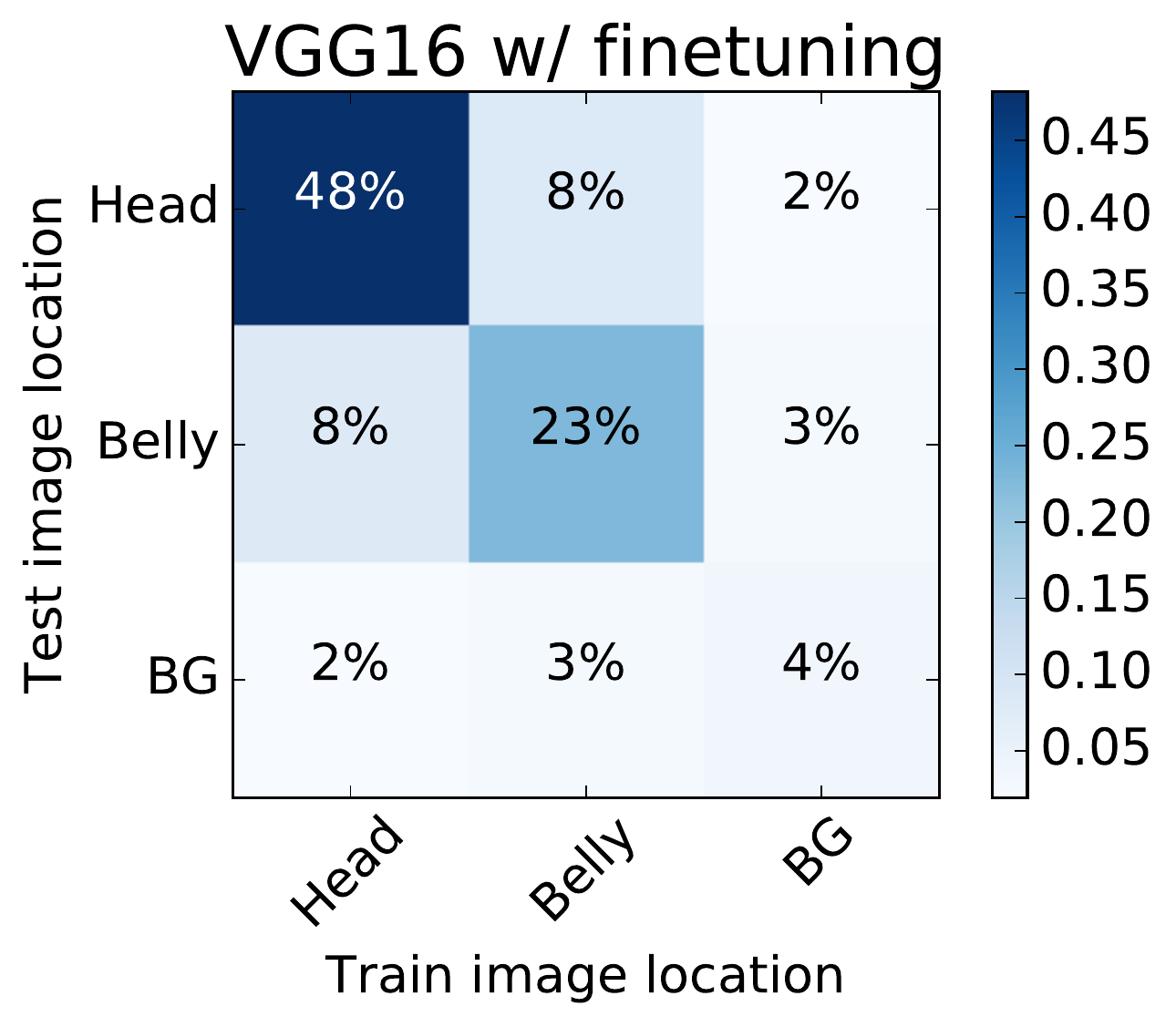}
 \caption{
 Influence of fine-tuning on the contribution of different bird body parts to the classification decision on CUB200-2011.
 $\alpha$ was fixed to a value of 2.0 for this experiment. 
 As can be seen, the bird's head gains influence at the cost of background areas. 
 $\alpha$ is set manually in this experiment.
 \label{fig:quantify-classification-vs-ft}}
\end{figure}

In addition to the influence of $\alpha$, we are also interested in the effect of fine-tuning on classification decisions. 
\figurename~\ref{fig:quantify-classification-vs-ft} depicts the results for the case of VGG16 and a fixed $\alpha=2$.
Fine-tuning seems to shift the focus more towards the bird's head, while especially the influence of background decreases.
This behavior is of course desired.
However, $\alpha$-pooling is one of the few approaches which allow quantifying the influence of different semantic object parts.

\section{Discussion\label{sec:discussion}}
\paragraph{Fine-grained tasks are about focusing on a few relevant areas}
Our in-depth analysis revealed that a high accuracy for fine-grained recognition can be achieved when only a few relevant areas are compared with
each other by implicit salient matching.
In terms of $\alpha$-pooling, this corresponds to a higher value of the parameter $\alpha$.
It also explains why bilinear pooling showed such a large performance gain for fine-grained recognition tasks~\cite{lin15bilinear}: the corresponding $\alpha=2$ causes an increase of influence for highly related features.
On the other hand, in generic image classification tasks like scene recognition, the general appearance seems more important and hence a lower value of $\alpha$ is better suited.
Our experiments showed that $\alpha=1.5$ is a good trade-off for a wide range of classification datasets and hence is a good starting point.
If fine-tuning is used, $\alpha$ will be learned and automatically adapted to the best value.

\paragraph{Implicit matching vs. explicit pose normalization}
The majority of approaches for fine-grained recognition assumes objects being decomposed into a few number
of parts~\cite{simon15nac,branson14posenorm,simon14discovery}.
It is common belief that part-based feature descriptors in contrast to global descriptors allow for a better representation for objects appearing in diverse poses.

In contrast, our analysis reveals that state-of-the-art global representations perform an implicit matching of several different image regions.
Compared to explicit part-based models, they are not limited by a fixed number of parts learned from the data or utilized during classification.
Our $\alpha$-pooling strategy can even learn how much a classification decision should rely on a few rather than a large number of matchings.
As argued in the last paragraph, the intuition that fine-grained recognition tasks are about ``detecting a small set of image regions that matter'' is right.
However, the consequence that explicit part-based models are the solution is questionable.
Rather than designing yet another part-based model, representations should be developed that lead to an even better implicit matching.

\begin{figure}
  \includegraphics[width=0.99\linewidth]{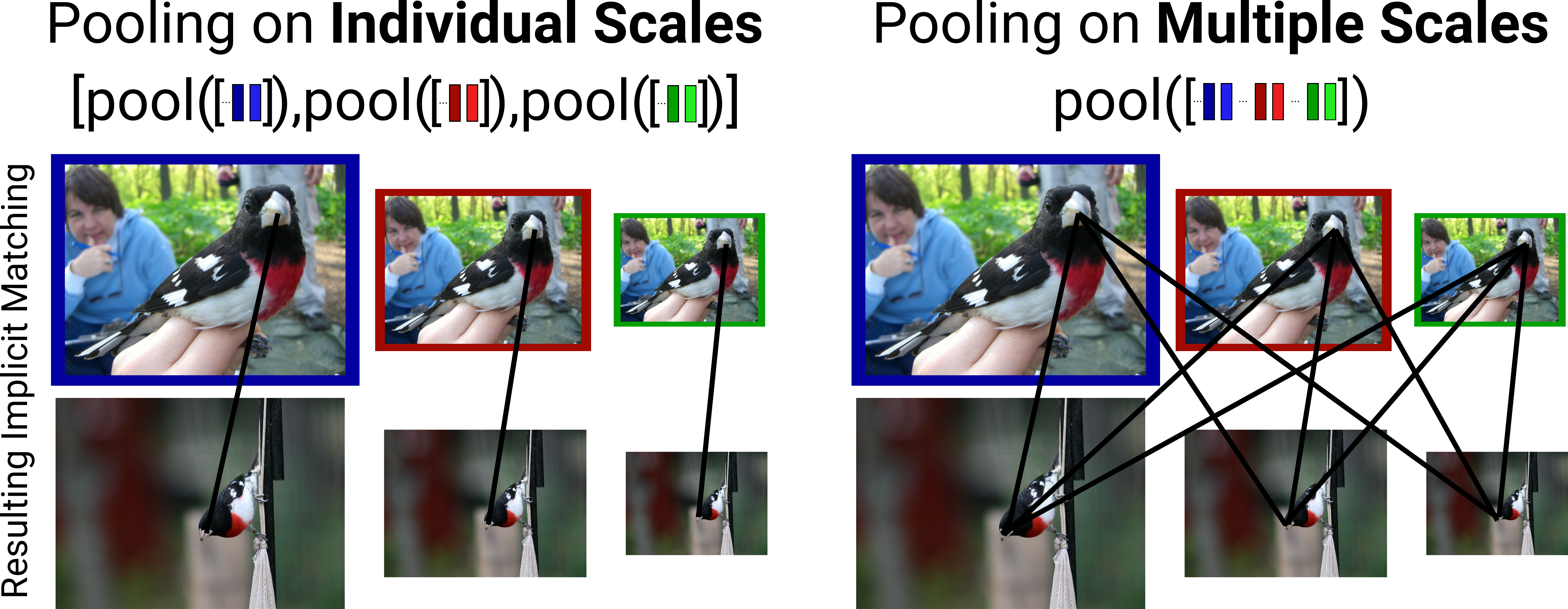}
  \caption{Illustration of different techniques to deal with multiple scales and their resulting implicit matching.
      Directly pooling over multiple scales allows for implicit matching across scales.}
  \label{fig:multiscale}
\end{figure}

\paragraph{Kernel view of classification decisions}
We argue that the kernel view of classification decisions is a valuable tool for understanding and analyzing different feature encoding.
We used the kernel view in the previous sections to show that a larger value of $\alpha$ focuses the classification decision on only a few most relevant pairs of local features.
This understanding also allowed us to visualize classification decision by using matchings to the most relevant training images.
However, there are even more possible ways to exploit this formulation in future work.
For example, we can derive a feature matching over multiple scales in a theoretically sound way.
In previous work, multiple scales were often handled by extracting crops at different scales and averaging the decision values of the last layer across all crops~\cite{he15resnet,szegedy16inception4}.
While this gives an improvement, a theoretical justification is missing.
In contrast, if we perform $\alpha$-pooling across all local features extracted from all scales of the input image, the kernel view reveals that this relates to a matching of local features across all possible combinations of locations and scales of two images (see \figurename~\ref{fig:multiscale}).
To summarize, while kernel functions are rarely explicitly used in state-of-the-art approaches, they can
be useful for both understanding and designing new approaches.

\vspace{-5pt}
\section{Conclusions}
\vspace{-5pt}
In this paper, we propose a novel generalization of average and bilinear pooling called $\alpha$-pooling.
Our approach has \textit{both} state-of-the-art performance and a clear justification of predictions.
It allows for a smooth transition between average and bilinear pooling, and to higher-order pooling, allowing for understanding the connection between these operating points. 
We find that in practice our method learns that an intermediate strategy between average and bilinear pooling offers the best performance on several fine-grained classification tasks. 
In addition, a novel way for visualizing classification predictions is presented showing the most influential training image regions for a decision.
Furthermore, we quantify the contributions of semantic parts in a classification decision based on these influential regions. 

\section*{Acknowledgments}
Part of this research was supported by grant RO 5093/1-1 of the German Research Foundation (DFG).
The authors thank Nvidia for GPU hardware donations.

{\small
\bibliographystyle{ieee}
\bibliography{paper}
}

\end{document}